\documentclass[letterpaper, 10 pt, conference]{ieeeconf}  

\IEEEoverridecommandlockouts                              
\usepackage[linesnumbered,ruled]{algorithm2e}
\usepackage{float}
\usepackage{subfigure}
\usepackage{graphics} 
\usepackage{epsfig} 
\usepackage{mathrsfs}
\usepackage{times} 
\usepackage{amsmath} 
\usepackage{amssymb}  
\usepackage{mathrsfs}
\usepackage{color}
\usepackage[table]{xcolor}
\usepackage[T1]{fontenc}
\usepackage{bm}
\usepackage{cite}
\usepackage{hyperref}
\usepackage{multirow}
\hypersetup{
	colorlinks=true,
	linkcolor=blue,
	filecolor=blue,      
	urlcolor=blue,
	citecolor=cyan,
}

\title{\LARGE \bf
 DR$^2$Track: Towards Real-Time Visual Tracking for UAV via \\ Distractor Repressed Dynamic Regression }
\author{Changhong Fu$^{1,*}$, Fangqiang Ding$^{1}$, Yiming Li$^{1}$, Jin Jin$^{1}$ and Chen Feng$^{2}$
	\thanks{*Corresponding author}
	\thanks{$^{1}$Changhong Fu, Fangqiang Ding, Yiming Li and Jin Jin are with the School of Mechanical Engineering, Tongji University, 201804 Shanghai, China.
		{\tt\small changhongfu@tongji.edu.cn}}
		\thanks{$^{2}$Chen Feng is with the Tandon School of Engineering, New York University, NY 11201 New York, United States.
		{\tt\small cfeng@nyu.edu}
	}%
}

\begin{document}
\maketitle
\thispagestyle{empty}
\pagestyle{empty}

\begin{abstract}
Visual tracking has yielded promising applications with unmanned aerial vehicle (UAV). In literature, the advanced discriminative correlation filter (DCF) type trackers generally distinguish the foreground from the background with a learned regressor which regresses the implicit circulated samples into a fixed target label. However, the predefined and unchanged regression target results in low robustness and adaptivity to uncertain aerial tracking scenarios. In this work, we exploit the local maximum points of the response map generated in the detection phase to automatically locate current distractors\footnote{In this paper, we refer the objects possessing a highly similar appearance with the tracked target as the distractors for conciseness.}. By repressing the response of distractors in the regressor learning, we can dynamically and adaptively alter our regression target to leverage the tracking robustness as well as adaptivity. Substantial experiments conducted on three challenging UAV benchmarks demonstrate both excellent performance and extraordinary speed ($\sim$50fps on a cheap CPU) of our tracker.
\end{abstract}
\section{INTRODUCTION}\label{sec:INTRODUCTION}
In recent years, empowering robots with object tracking capability has witnessed an incremental number of real-world applications, \emph{e.g.}, path planning~\cite{laguna2019path}, aerial cinematography~\cite{bonatti2019towards}, and collision avoidance~\cite{fu2014robust}, among which tracking object with unmanned aerial vehicle (UAV) is intensively studied due to the high mobility, fast speed, and credible portability of UAV. However, UAV object tracking remains a tough task due to the complicated aerial scenarios, \emph{e.g.}, background distractors, UAV/object fast motion, and intrinsic defects of UAV such as harsh computation resources, restricted power supplement as well as severe vibration. 

Discriminative correlation filter (DCF) type methods ~\cite{henriques2015high,li2018learning,li2020autotrack} and deep neural network (DNN)-based approaches~\cite{bertinetto2016fully,valmadre2017end,guo2017learning} have become two main research topics in the visual tracking community. Though achieving appealing performance, DNN-based trackers mostly suffer from heavy computation burden, which conflicts with limited computation resources and power capacity of UAV. To this end, DCF type methods are employed in our work for UAV object tracking owing to its high calculation efficiency and satisfactory performance. The core idea of DCF type approaches is that cyclic correlation or convolution in the spatial domain is able to be calculated via element-wise multiplication in the Fourier domain. Hence, training samples can be implicitly 
\begin{figure}[t]
	\includegraphics[width=1\columnwidth]{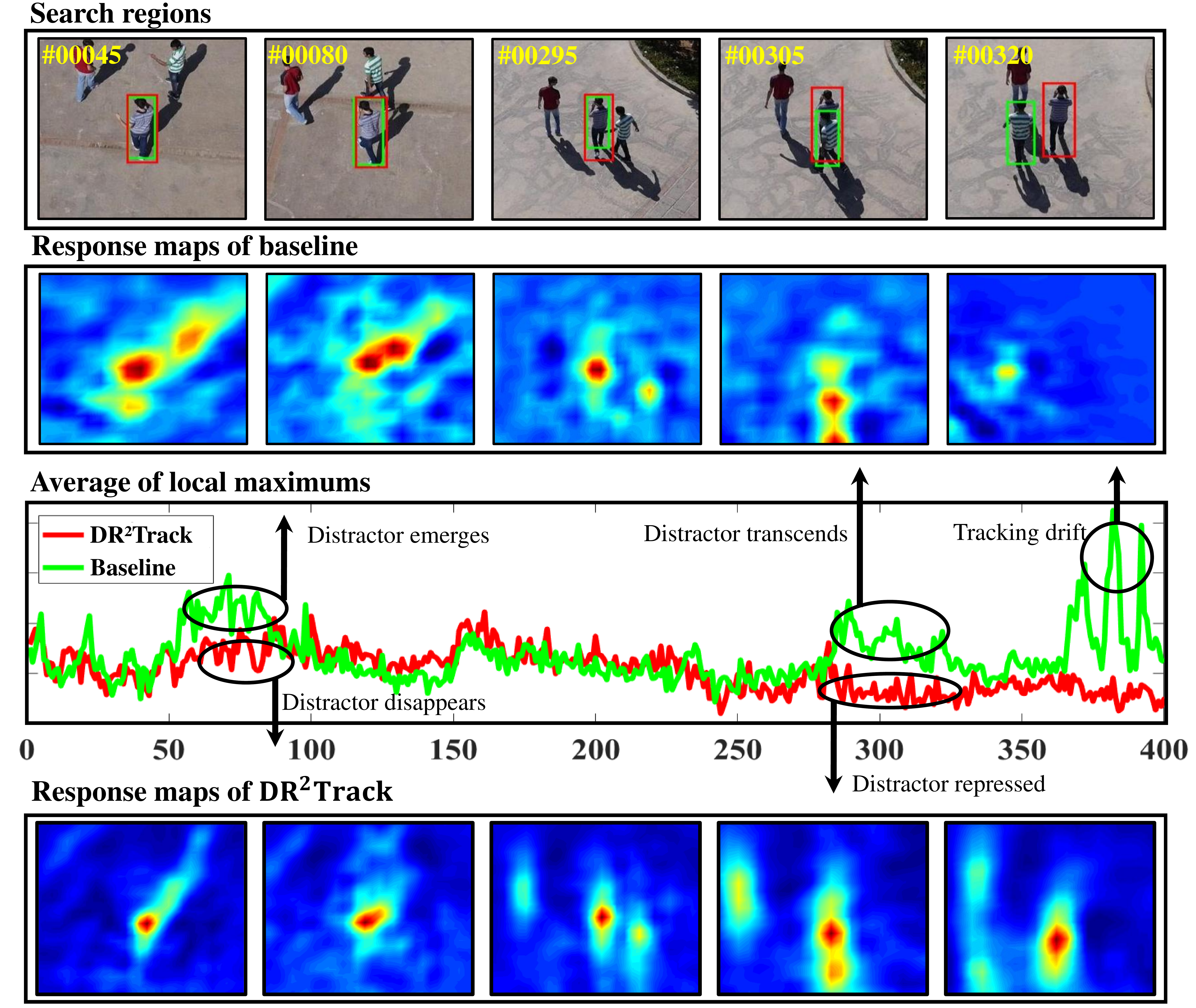}
	\caption{Illustration of DR$^2$Track's effectiveness in distractor repression. \textcolor[rgb]{1,0,0}{\textbf{Red}} and \textcolor[rgb]{0,1,0}{\textbf{green}} boxes respectively represent the output of ours and baseline\cite{li2018learning}. Compared to the response maps of \cite{li2018learning}, ours is more concentrated. The averages of top $N$ ($N=30$) background local maximums are also exhibited. When the distractor comes close to the object in frame $\#80$, the distractor does not emerge in our response maps due to the successive suppression of local maximums. In frame $\#305$ of\cite{li2018learning}, the response of distractor transcends that of the object, resulting in tracking drift.}
	\label{fig:scene}

\end{figure}
augmented by cyclic translations of the original sample, which upgrades the discrimination of the filter without losing speed. 

In literature, the work-flow of DCF type trackers is two-fold: 1) training phase: a regression objective is minimized in the Fourier domain to calculate an optimal filter, 2) detection phase: the object is localized by searching for the maximum score in the response map obtained by correlation between the filter and feature map. Ideally, the response map is unimodal and has similar shape to the fixed Gaussian-shaped regression label. However, in practice, the generated response map usually has multiple peaks, because background distractors can possess large correlation response with the online-learned filter, as shown in Fig.~\ref{fig:scene}. In the worst situation, tracker will drift to the distractor with the largest peak value. Based on the aforementioned observation, this work fully uses the local maximums points in the response map to locate the distractors in a dynamic and adaptive manner. After that, we enforce a pixel-level penalization on the distractors by reducing their label values to repress their interferences. 

The presented tracker, named {\textbf{D}istractor \textbf{R}epressed \textbf{D}ynamic \textbf{R}egression Tracker} (DR$^2$Track), performs outstandingly with only hand-crafted features and works at around 50 fps on a cheap CPU. Most tellingly, DR$^2$Track has comparable performance and even exceeds existing state-of-the-art DNN-based trackers. Additionally, in consideration of the effective repression of background distractors, the search area can be properly enlarged to enrich the negative samples, at the same time enhancing the robustness against fast motion. The main novelties of this paper are three-fold:
\begin{itemize}
	\item A dynamic rather than constant regression problem is learned in this work for efficient and effective visual object tracking onboard UAVs.
	\item  Local maximum points in the response map are made full use of to automatically and adaptively discover as well as repress background distractors on the fly.
	\item Substantial experiments are conducted on three demanding UAV datasets to prove the efficacy and efficiency of DR$^2$Track in the real-world UAV tracking scenarios.
\end{itemize}

\section{RELATED WORKS}\label{sec:RELATEDWORK}

\subsection{Discriminative methods for visual tracking}
In visual tracking, discriminative methods learn a regressor online for background-foreground classification with both negative and positive samples, while generative methods commonly focus on modeling the object appearance.
In \cite{grabner2006real}, an online boosting tracker is proposed by merging multiple weak classifiers. B. Babenko \emph{et al.}~\cite{babenko2011robust} introduced multiple instances learning to significantly raise the tracking performance. S. Hare \emph{et al.}~\cite{hare2016struck} trained a kernelized structured classifier to track objects adaptively. Beyond traditional online-training discriminative approaches, some end-to-end deep frameworks \cite{bertinetto2016fully,valmadre2017end,guo2017learning} apply offline-learning models for visual object tracking. As a branch of discriminative methods, DCF type trackers~\cite{henriques2015high,danelljan2017eco,li2018learning,Li2020ICRA} have attracted extensive research attention owing to their computational efficiency and prominent tracking performance.
\subsection{Discriminative correlation filter type approaches}
The pioneer tracker, proposed by D. S. Bolme \cite{bolme2010visual}, trained filters by minimizing the sum of squared error between the actual and predefined label, and exhibited extraordinary efficiency. Since \cite{bolme2010visual}, DCF type trackers have been intensively investigated to boost the tracking performance: object representation augmentation \cite{henriques2015high,danelljan2014adaptive,ma2015hierarchical}, spatial regularization \cite{danelljan2015learning,dai2019visual}, training sample management \cite{danelljan2016adaptive}, continuous convolution \cite{danelljan2016beyond}, and multiple scales \cite{danelljan2017discriminative, li2014scale}. However, the existing methods commonly adopt a fixed Gaussian-shaped label as the target response in the regressor learning, which can only learn a rough classification boundary and cannot be generalized to various unpredictable UAV tracking scenarios.
\subsection{Prior works in distractor repression}
M. Muller \emph{et al.}~\cite{mueller2017context} fed context patches located around the tracked object into the objective function to cope with background clutter. Yet it cannot automatically detect the background distraction. Z. Huang \emph{et al.}~\cite{huang2019learning} proposed to regulate the variation of response map in the training phase for aberrance repression. However, this technique is sensitive to the regularization term: an overly low penalty has no influence while an overly high one leads to overfitting issues. Z. Zhu \emph{et al.} \cite{zhu2018distractor} employed a novel sampling strategy to learn distractor-aware features in offline training and designed a distractor-aware module online. Nevertheless, it consumes much computation resources for distractor suppression. Different from the above trackers, DR$^2$Track can efficiently repress the adaptively located distractors by accordingly adjusting the target label in the regression equation.
\subsection{UAV-oriented tracking methods}
In UAV tracking scenarios, more difficulties are introduced than the generic object tracking based on the stationary camera. Furthermore, the mobile UAV platform has extremely limited computation resources assigned to visual tracking. To this end, object tracker possessing strong robustness and fast running speed is hence desirable.  In~\cite{fu2014robust} and \cite{cheng2017an}, the importance of test sample and re-detection algorithm were integrated to promote the tracking robustness against challenging aerial attributes. \cite{li2020keyfilter,li2020intermittent} introduced keyfilter technique into DCF to mitigate filter degradation by strong temporal restriction. However, the real-time requirement is not satisfied. In contrast, DR$^2$Track has outstanding tracking performance and occupies little computation resources without casting aside real-time speed. 
\begin{figure*}[!t]
	\centering
	\includegraphics[width=1\textwidth]{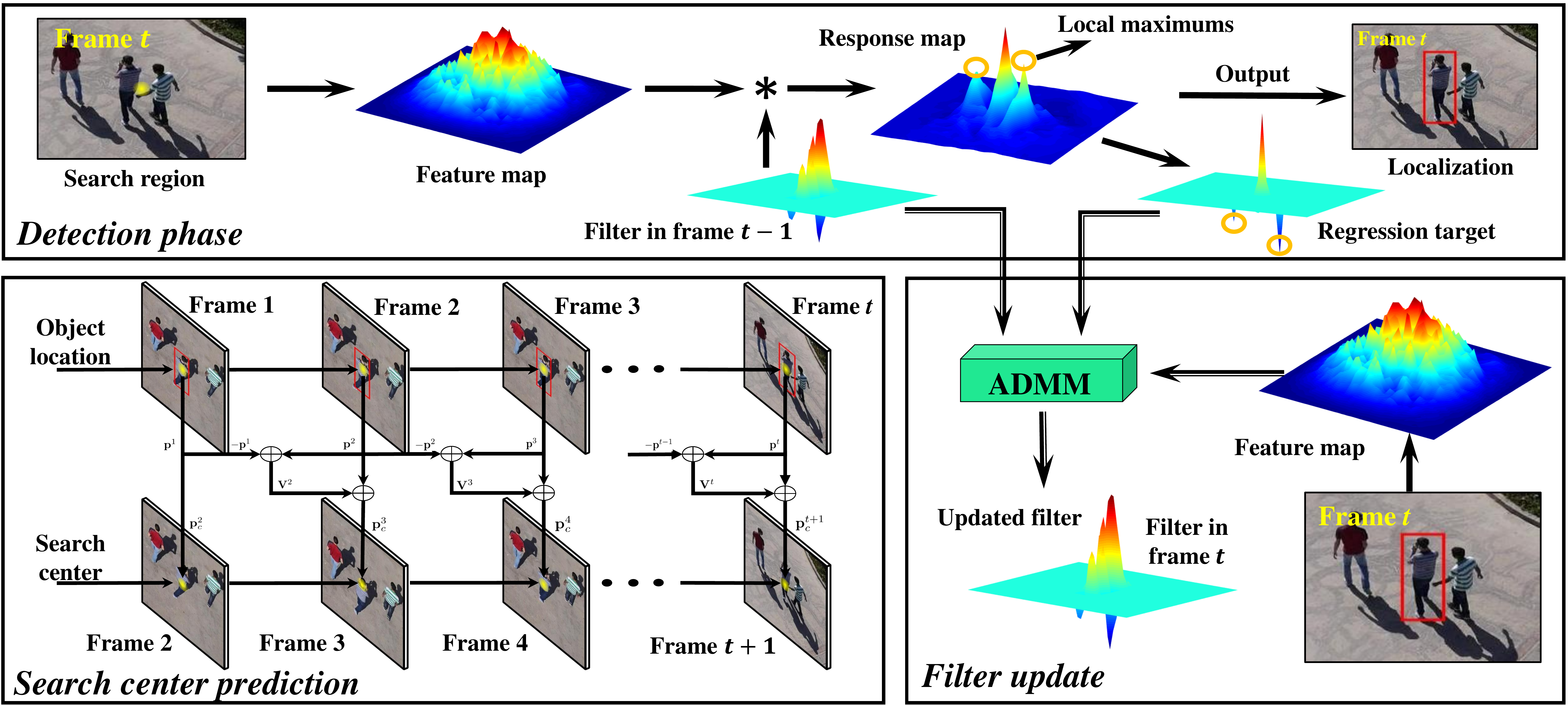}
	\caption{Overall work-flow of DR$^2$Track. In the detection phase, the search region is sampled around the predicted center location displayed as a yellow point in the figure. After the feature extraction, the response map is obtained by the correlation between the filter and feature map. In the end, the current location is gained by searching for the maximum value of the response map while the search area in the next frame is updated. In the training phase, the local maximums of the response map are repressed via pixel-level penalization in the regression target, ADMM is used to calculate the optimal filter.}
	\label{fig:mainstructure}
\end{figure*}
\section{PROPOSED METHOD}\label{sec:METHOD}
\subsection{Review spatial-temporal regularized correlation filters}
This work uses spatial-temporal regularized correlation filters (STRCF) \cite{li2018learning} as the baseline due to its satisfactory performance stemming from spatial-temporal regularization. The minimized objective can be written as:
		\begin{equation}\small\label{eqn:1}
		E(\mathbf{h})=	\frac{1}{2}\Vert\mathbf{g}-\sum_{c=1}^{C}\mathbf{m}_c\ast\mathbf{h}_c\Vert_2^2+\sum_{c=1}^{C}\Big(\frac{\theta}{2}\Vert\mathbf{h}_c-{\mathbf{h}}^l_c\Vert_2^2+\frac{1}{2}\Vert\mathbf{w}\odot\mathbf{h}_c\Vert_2^2\Big) \ ,
		\end{equation}
	\noindent where $\mathbf{g}\in \mathbb{R}^{K\times 1}$ represents fixed Gaussian-shape regression target with length $K$, $\mathbf{m}_c\in \mathbb{R}^{K\times 1}$ denotes the vectorized feature map of training patch in $c$-th channel and $\mathbf{h}=[\mathbf{h}_1,\mathbf{h}_2,\dots,\mathbf{h}_C]\in \mathbb{R}^{K\times C}$ represents the filters in all feature channels. Besides, ${\mathbf{h}^{l}}$ denotes the filter trained in the last frame, $*$ is the cyclic convolution operator, and $\odot$ denotes the element-wise multiplication. $\mathbf{w}\in \mathbb{R}^{K\times 1}$ is spatial regularization term for boundary effect mitigation and $\theta$ represents temporal regularization term for temporal restriction.
\subsection{Overall formulation of DR$^2$Track}
		Contrary to STRCF with a fixed target label $\mathbf{g}$, we introduce a distractor-repressed vector $\mathbf{d}$ to generate a dynamic regression target. Our objective is to minimize:
		\begin{equation}\footnotesize\label{eqn:2}
		E(\mathbf{h})=\frac{1}{2}\Vert\mathbf{g}\odot \mathbf{d}-\sum_{c=1}^{C}\mathbf{m}_c\ast\mathbf{h}_c\Vert_2^2+\sum_{c=1}^{C}\frac{1}{2}\Vert\mathbf{w}\odot\mathbf{h}_c\Vert_2^2+\sum_{c=1}^{C}\frac{\theta}{2}\Vert\mathbf{h}_c-{\mathbf{h}}^l_c\Vert_2^2  .
		\end{equation}
		 
		 As shown in Fig. \ref{fig:scene}, the local maximum points in response map $\mathbf{R}$ (defined in Eq. (\ref{eqn:12})) indicate the presence of distrators. Hence, they are  integrated into the formulation of $\mathbf{d}$. The distractor-repressed vector can be obtained by:
		\begin{equation}\small\label{eqn:3}
			\mathbf{d}=\mathbf{I}-\mu \mathbf{P}^{\top}\Delta(\mathbf{R[\eta]}) \ ,
		\end{equation} 
		 \noindent where $\Delta(\cdot)$ represents local-maximum-cropping function. Generally, there are many local maximum points in the background. However, most of them cannot be counted as distractors for their low response values. Thus, in terms of implementation, top $N$ local maximums are selected. $\mathbf{P}^{\top}$ cuts the central area of $\Delta(\cdot)$ in order to remove the maximum points within the object area and $\mu$ is a factor controlling the repression strength. $\mathbf{I}$ denotes an identity vector. $[\eta]$ denotes a shift operator to match peaks of response map and regression target. The overall work-flow is illustrated in Fig.~\ref{fig:mainstructure}.
		\subsection{Optimization of formulation}
		Due to the convexity of Eq.~(\ref{eqn:2}), the globally optimal solution can be obtained via iteratively using the alternating direction method of multipliers (ADMM) algorithm. Defining auxiliary
		variable $\mathbf{v}=\mathbf{h} (\mathbf{v}=[\mathbf{v}_1,\mathbf{v}_2,\dots,\mathbf{v}_C])$, the augmented Lagrangian form of Eq.~(\ref{eqn:2}) is formulated by:
		\begin{equation}\small\label{eqn:4}
		\begin{aligned}
		&L(\mathbf{h},\mathbf{v},\mathbf{u})=\frac{1}{2}\Vert(\mathbf{g}\odot \mathbf{d})-\sum_{c=1}^{C}\mathbf{m}_c*\mathbf{v}_c\Vert_2^2+\frac{\theta}{2}\sum_{c=1}^{C}\Vert\mathbf{v}_c-\mathbf{v}_c^{l}\Vert_2^2 \\
		&+\frac{1}{2}\sum_{c=1}^{C}\Vert\mathbf{w}\odot\mathbf{h}_c\Vert_2^2 +\sum_{c=1}^{C}\mathbf{u}_c^{\top}(\mathbf{v}_c-\mathbf{h}_c)+\frac{\gamma}{2}\sum_{c=1}^{C}\Vert\mathbf{v}_c-\mathbf{h}_c\Vert^2_2
		\end{aligned}\ .
	 	\end{equation}
		
		Here, $\mathbf{u}=[\mathbf{u}_1,\mathbf{u}_2,\dots,\mathbf{u}_C]$ represents the Lagrange multiplier and $\gamma$ denotes the step size parameter. $\mathbf{z}=\mathbf{u}/{\gamma}$ is introduced to simplify Eq.~(\ref{eqn:4}), which is rewritten as:
		\begin{equation}\small\label{eqn:5}
		\begin{aligned}
		L(\mathbf{h},\mathbf{v},\mathbf{z})&=\frac{1}{2}\Vert(\mathbf{g}\odot \mathbf{d})-\sum_{c=1}^{C}\mathbf{m}_c*\mathbf{v}_c\Vert_2^2+\frac{\theta}{2}\sum_{c=1}^{C}\Vert\mathbf{v}_c-\mathbf{v}_c^{l}\Vert_2^2 \\ &+\frac{1}{2}\sum_{c=1}^{C}\Vert\mathbf{w}\odot\mathbf{h}_c\Vert_2^2+\frac{\gamma}{2}\sum_{c=1}^{C}\Vert\mathbf{v}_c-\mathbf{h}_c+\mathbf{z}_c\Vert_2^2 
		\end{aligned}\ ,
		\end{equation}
		\noindent where $\mathbf{z}=[\mathbf{z}_1,\mathbf{z}_2,\dots,\mathbf{z}_C]\in \mathbb{R}^{K\times C}$. Then the ADMM algorithm is utilized via alternatively solving three subproblems, \emph{i.e.}, $\mathbf{h}$, $\mathbf{v}$, and $\mathbf{z}$, for $E$ iterations.
		
		\begin{subsubsection}{\textbf{Solution to subproblem} $\mathbf{v}$}
Using $\mathbf{h}$ and $\mathbf{z}$ obtained in the last iteration, the optimal $\widehat{\mathbf{v}}$ in the $e$-th ($e=1,2,\cdots,E$) iteration can be determined by:
\begin{equation}
			\small
			\label{eqn:6}
			\begin{aligned}
			\mathbf{v}_c^e&=\mathop{\arg\min}\limits_{\mathbf{v}_c}\frac{1}{2}\Vert(\mathbf{g}\odot \mathbf{d})-\sum_{c=1}^{C}\mathbf{m}_c*\mathbf{v}_c\Vert_2^2+\frac{\theta}{2}\sum_{c=1}^{C}\Vert\mathbf{v}_c-\mathbf{v}_c^{l}\Vert_2^2\\&+\frac{\gamma}{2}\sum_{c=1}^{C}\Vert\mathbf{v}_c-\mathbf{h}_c^{e-1}+\mathbf{z}_c^{e-1}\Vert_2^2
			\end{aligned}	\ . 
\end{equation}	
			\noindent\textbf{\emph{Remark 1}:} In the first iteration of the ADMM algorithm, zero matrix acts as $\mathbf{h}^{e-1}$ and $\mathbf{z}^{e-1}$ for the optimization in Eq.~(\ref{eqn:6}). For the first frame, $\theta$ is set as 0 in the training phase.
			
			Based on the convolution theorem, the cyclic convolution operation in spatial domain denoted by $*$ can be replaced by element-wise multiplication $\odot$ in the Fourier domain. The Fourier form of Eq.~(\ref{eqn:6}) can be written as follows:
			\begin{equation}\small\label{eqn.7}
			\begin{aligned}
			\widehat{\mathbf{v}}_c^e&=\mathop{\arg\min}\limits_{\widehat{\mathbf{v}}_c}\frac{1}{2}\Vert(\widehat{\mathbf{g}\odot \mathbf{d}})-\sum_{c=1}^{C}\widehat{\mathbf{m}}_c\odot\widehat{\mathbf{v}}_c\Vert_2^2+\frac{\theta}{2}\sum_{c=1}^{C}\Vert\widehat{\mathbf{v}}_c-\widehat{\mathbf{v}}_c^{l}\Vert_2^2\\&+\frac{\gamma}{2}\sum_{c=1}^{C}\Vert\widehat{\mathbf{v}}_c-\widehat{\mathbf{h}}_c^{e-1}+\widehat{\mathbf{z}}_c^{e-1}\Vert_2^2
			\end{aligned}\ ,
			\end{equation}
			\noindent where $\widehat{\cdot}$ denotes the Fourier transformation. In consideration of the independence of each pixel, the solution can be obtained respectively across all channels for every pixel. The optimization in the $j$-th pixel can be reformulated as:
			\begin{equation}\small\label{eqn:8}
			\begin{aligned}
			&\Psi_j(\widehat{\mathbf{v}}^e)=\mathop{\arg\min}\limits_{\Psi_j(\widehat{\mathbf{v}})}\frac{1}{2}\Vert({\widehat{\mathbf{g}\odot \mathbf{d}}})_j-\Psi_j(\widehat{\mathbf{m}})^{\top}\Psi_j(\widehat{\mathbf{v}})\Vert_2^2\\ &+\frac{\theta}{2}\Vert\Psi_j(\widehat{\mathbf{v}})-\Psi_j(\widehat{\mathbf{v}}^{l})\Vert_2^2+\frac{\gamma}{2}\Vert\Psi_j(\widehat{\mathbf{v}})-\Psi_j(\widehat{\mathbf{h}}^{e-1})+\Psi_j(\widehat{\mathbf{z}}^{e-1})\Vert
			\end{aligned}\ ,
			\end{equation} 
			\noindent where $\Psi_j(\cdot)\in \mathbb{R}^{1\times C}$\ denotes the values across all channels on pixel $j$. Applying Sherman-Morrison formula when solving Eq.~(\ref{eqn:8}), the analytical solution can be deviated to:
			\begin{equation}
			\begin{aligned}\small\label{eqn:9}
			&\Psi_j(\widehat{\mathbf{v}}^e)=\frac{1}{\theta+\gamma}(\mathbf{I}-\frac{\Psi_j(\widehat{\mathbf{m}})\Psi_j(\widehat{\mathbf{m}})^{\top}}{\theta+\gamma+\Psi_j(\widehat{\mathbf{m}})^{\top}\Psi_j(\widehat{\mathbf{m}})})\\ 
			&(\Psi_j(\widehat{\mathbf{m}})({\widehat{\mathbf{g}\odot \mathbf{d}}})_j+\theta\Psi_j(\widehat{\mathbf{v}}^{l})+\gamma\Psi_j(\widehat{\mathbf{h}}^{e-1})-\gamma\Psi_j(\widehat{\mathbf{z}}^{e-1})) 
			\end{aligned}
			\ .
			\end{equation}
		\end{subsubsection}
	\begin{subsubsection}{\textbf{Solution to subproblem} $\mathbf{h}$}
	 In the $e$-th iteration, given $\mathbf{v}^e$ and $\mathbf{z}^{e-1}$, the optimal $\mathbf{h}^e$ ($e=1,2,\cdots,E$) can be calculated as:
	\begin{equation}\small\label{eqn:10}
	\begin{aligned}
	\mathbf{h}_c^e=&\mathop{\arg\min}\limits_{\mathbf{h}_c}\frac{1}{2}\sum_{c=1}^{C}\Vert\mathbf{w}\odot\mathbf{h}_c\Vert_2^2+\frac{\gamma}{2}\sum_{c=1}^{C}\Vert\mathbf{v}^{e}_c-\mathbf{h}_c+\mathbf{z}^{e-1}_c\Vert_2^2 \\
	=&(\mathbf{W}^{\top}\mathbf{W}+\gamma K\mathbf{I})^{-1}\gamma K(\mathbf{v}^{e}_c+\mathbf{z}^{e-1}_c)
	=\frac{\gamma K(\mathbf{v}^{e}_c+\mathbf{z}^{e-1}_c)}{(\mathbf{w}\odot\mathbf{w})+\gamma K}
	\end{aligned}~,
	\end{equation}
	\noindent where $\mathbf{W}=diag(\mathbf{w})\in \mathbb{R}^{K\times K}$ denotes the diagonal matrix.
	\end{subsubsection}

\begin{subsubsection}{\textbf{Update of Lagrangian Multiplier $\mathbf{z}$}}
 After solving $\mathbf{v}^{e}$ and $\mathbf{h}^{e}$ in the current iteration, the Lagrangian multipliers are updated by:
\begin{equation}\small\label{eqn:11}
	\mathbf{z}^{e}=\mathbf{z}^{e-1}+\gamma^{e-1}(\mathbf{v}^{e}-\mathbf{h}^{e})\ ,
\end{equation}
where $\gamma^{e}=\mathop{\min}(\gamma_{max},\beta \gamma^{e-1})$ is reset after every iteration.
\end{subsubsection}
\begin{subsection}{Object localization}
	When a new frame arrives, the filter trained in the last frame $\mathbf{h}^{t-1}$ is used to localize the object through searching for the peak in the response map calculated by the following formula:
	\begin{equation}\small\label{eqn:12}
		\mathbf{R}=\mathcal{F}^{-1}\Big(\sum_{c=1}^{C}\widehat{\mathbf{s}}_c^{t}\odot\widehat{\mathbf{h}}_c^{t-1}\Big) \ ,
	\end{equation}
	\noindent where $\mathbf{s}_c^{t}\in \mathbb{R}^{K\times 1}$ denotes the feature map of the search area patch in the $c$-th channel and $\mathcal{F}^{-1}$ represents the inverse diverse Fourier transform. For normalization, the response map is divided by its maximum value. Traditionally, in frame $t$, CF-based trackers extract the search area patch centering at $\mathbf{p}^{t-1}=[x^{t-1},y^{t-1}]$ which denotes the estimated location in frame $t-1$. Here, $x$ and $y$ respectively denotes the horizontal and vertical coordinate of object. In this work, we also use a simple but effective strategy to localize the search area. Specifically, object's moving speed within the last two frames is utilized to coarsely predict the current object's location.  Denoted by $\mathbf{V}^t=[V_x^t,V_y^t]$, the velocity vector in frame $t$ is obtained by:
	\begin{equation}\small\label{eqn:13}
		[V_x^{t},V_y^{t}]=[x^t,y^t]-[x^{t-1},y^{t-1}] \ ,
	\end{equation} 
	\noindent then the center of search area in frame $t+1$, \emph{i.e.}, $\mathbf{p}^{t+1}_c$, is decided by:
	\begin{equation}\small\label{eqn:14}	
	[x^{t+1}_c,y^{t+1}_c]=[x^{t},y^{t}]+[V_x^t,V_y^t]~.
	\end{equation}
\end{subsection}
\begin{algorithm}[!h]
	\caption{DR$^2$Track}
	\KwIn{Groundtruth in the first frame\\
	\hspace{1.0cm}	Subsequent images in the stream
	}	
	\KwOut{Estimated object location in frame $t$}
	\label{alg:workflow}
	\eIf{$t=1$}{
		Extract $\mathbf{m}^1$ centered at the groundtruth $\mathbf{p}^1$\\
		Use Eq.~(\ref{eqn:9}-\ref{eqn:11}) to initialize the filters $\mathbf{h}^1$ and $\mathbf{v}^l$\\
		Initialize object velocity $\mathbf{V}^1=\mathbf{0}$ 
	}
	{
		Obtain the search center $\mathbf{p}_c^t$ by Eq.~(\ref{eqn:14})\\
		Extract $\mathbf{s}^t$ centered at $\mathbf{p}_c^t$\\
		Use Eq.~(\ref{eqn:12}) to generate the response map $\mathbf{R}$\\
		Find the peak location $\mathbf{p}^t$ of $\mathbf{R}$ and output\\
		Extract $\mathbf{m}^{t}$ centered at $\mathbf{p}^t$\\
		Use Eq.~(\ref{eqn:9}-\ref{eqn:11}) to update the filters $\mathbf{h}^t$ and $\mathbf{v}^l$\\
		Update object velocity $\mathbf{V}^t$ by Eq.~(\ref{eqn:13}).
	}	
\end{algorithm}
\section{EXPERIMENTS}\label{sec:EXPERIMENT}
To comprehensively and rigorously assess our tracking performance in the real-word UAV tracking task, three canonical and stringent UAV benchmarks, \emph{i.e.}, UAV123@10fps\cite{Mueller2016ECCV}, DTB70\cite{Li2017AAAI} and UAVDT~\cite{Du2018ECCV}, are employed 
	\begin{figure*}[!t]
	\begin{center}
		\subfigure { \label{fig:DTB70} 
			\begin{minipage}{0.315\textwidth}
				\centering
				\includegraphics[width=1\columnwidth]{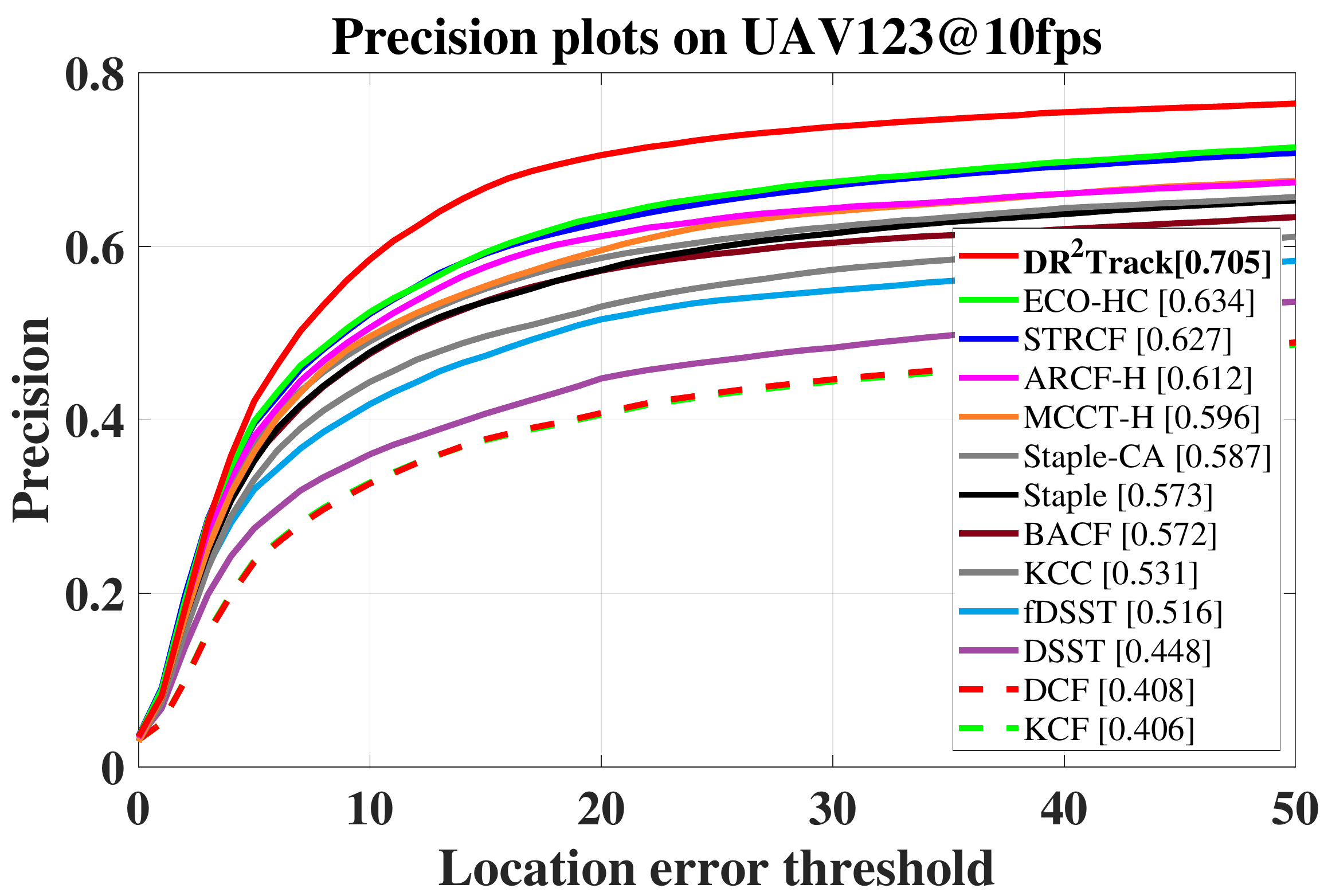}
				\\
				\includegraphics[width=1\columnwidth]{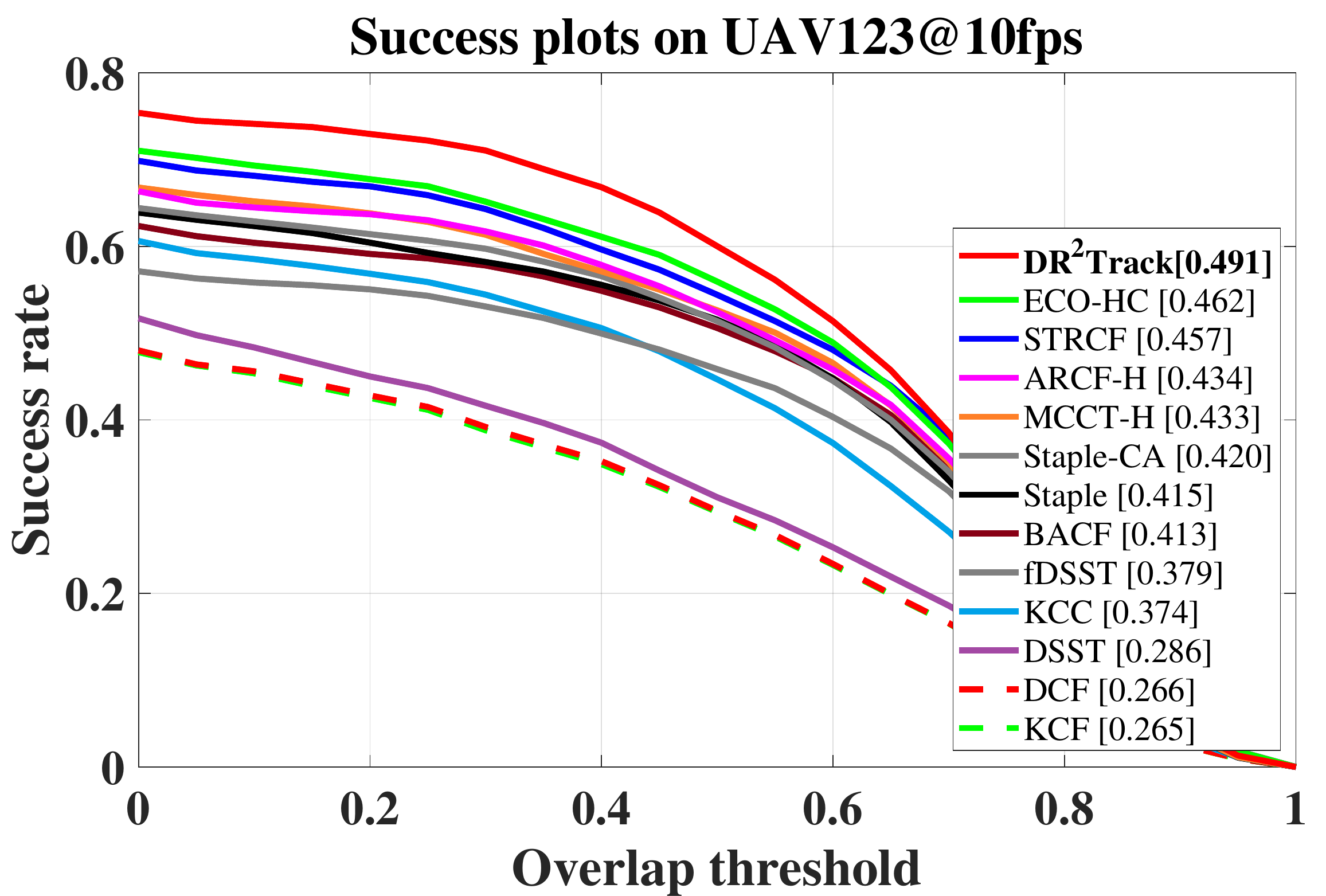}
			\end{minipage}
		}
		\subfigure { \label{fig:UAVDT} 
			\begin{minipage}{0.315\textwidth}
				\centering
				\includegraphics[width=1\columnwidth]{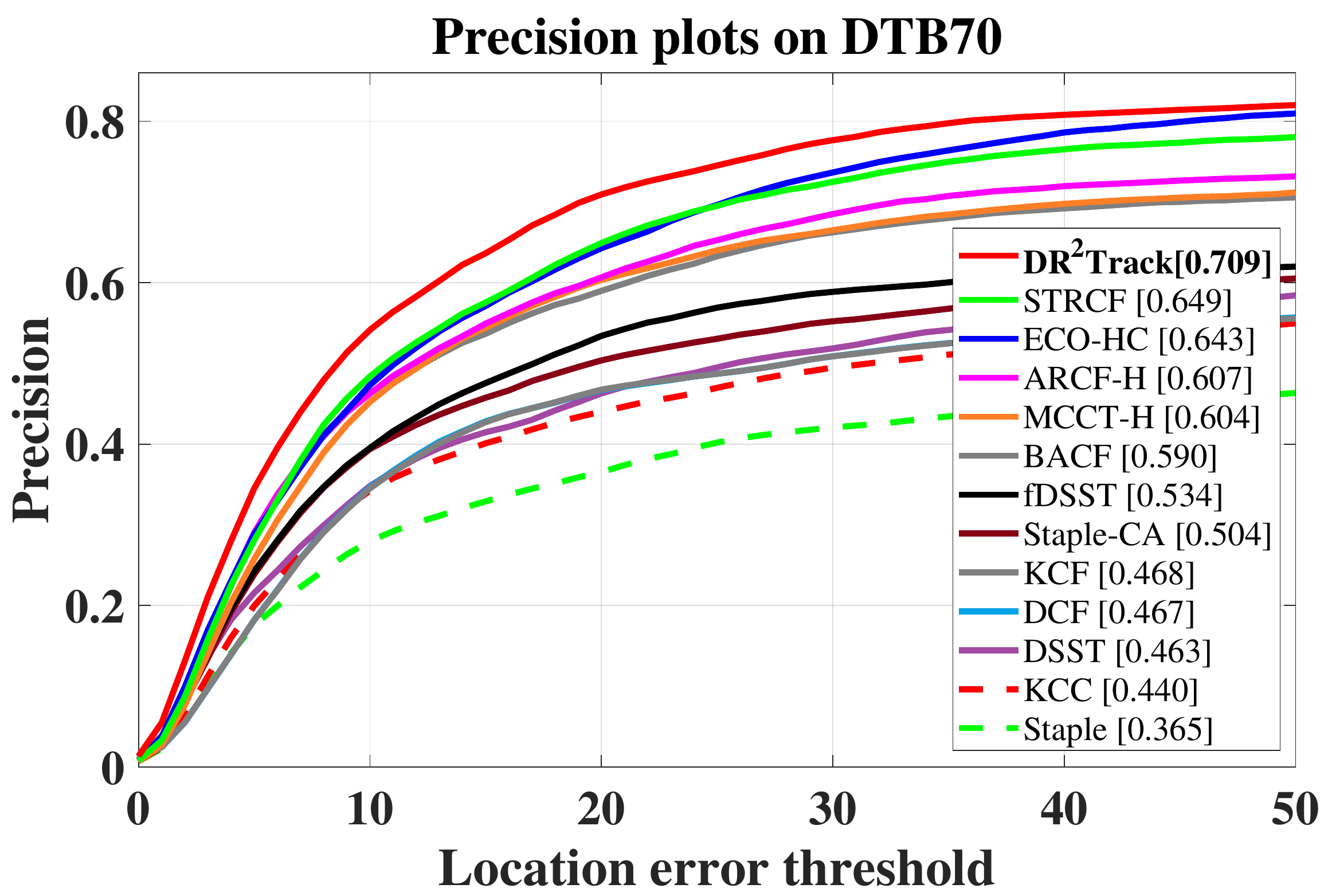}
				\\
				\includegraphics[width=1\columnwidth]{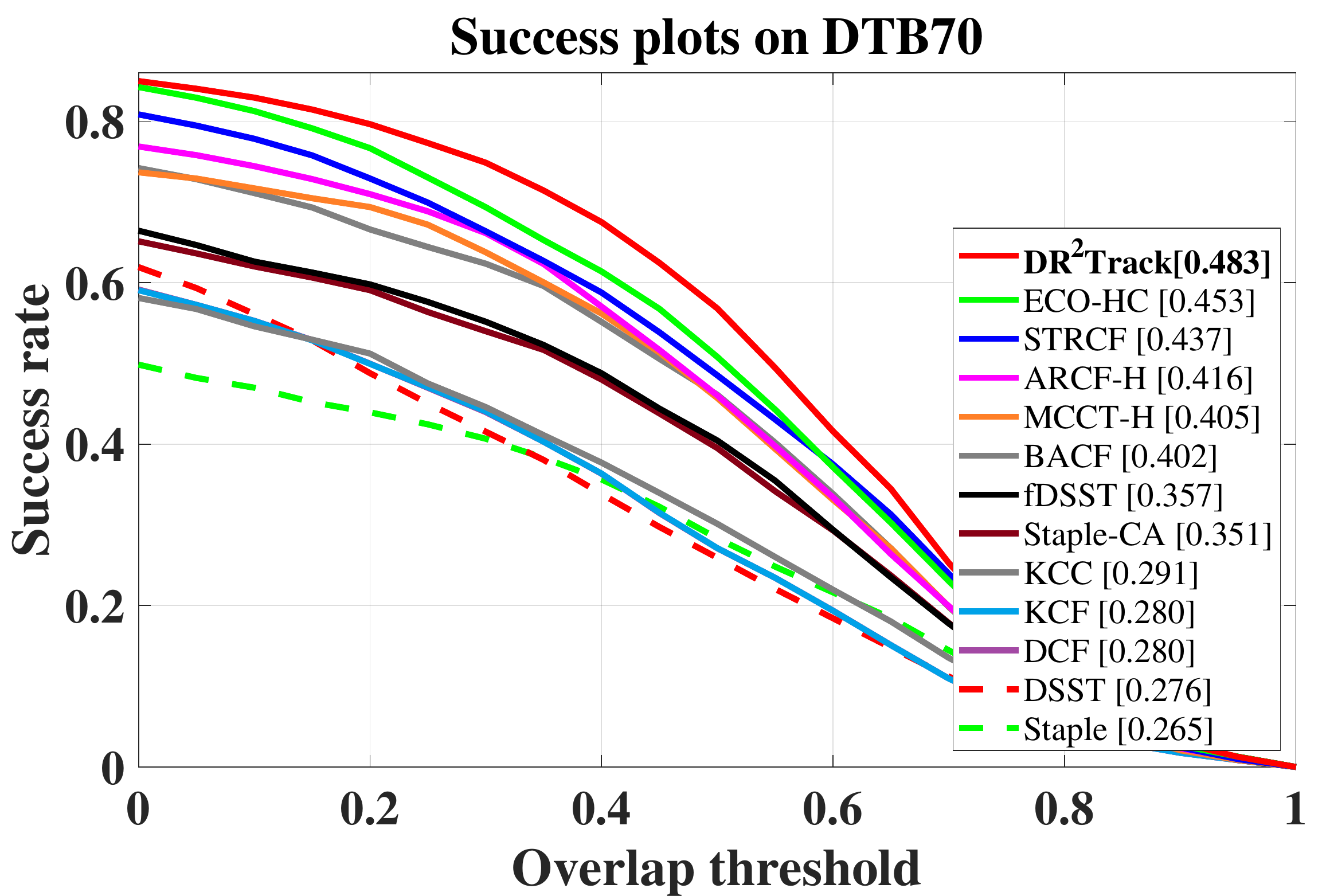}
			\end{minipage}
		}
		\subfigure { \label{fig:UAV123} 
			\begin{minipage}{0.315\textwidth}
				\centering
				\includegraphics[width=1\columnwidth]{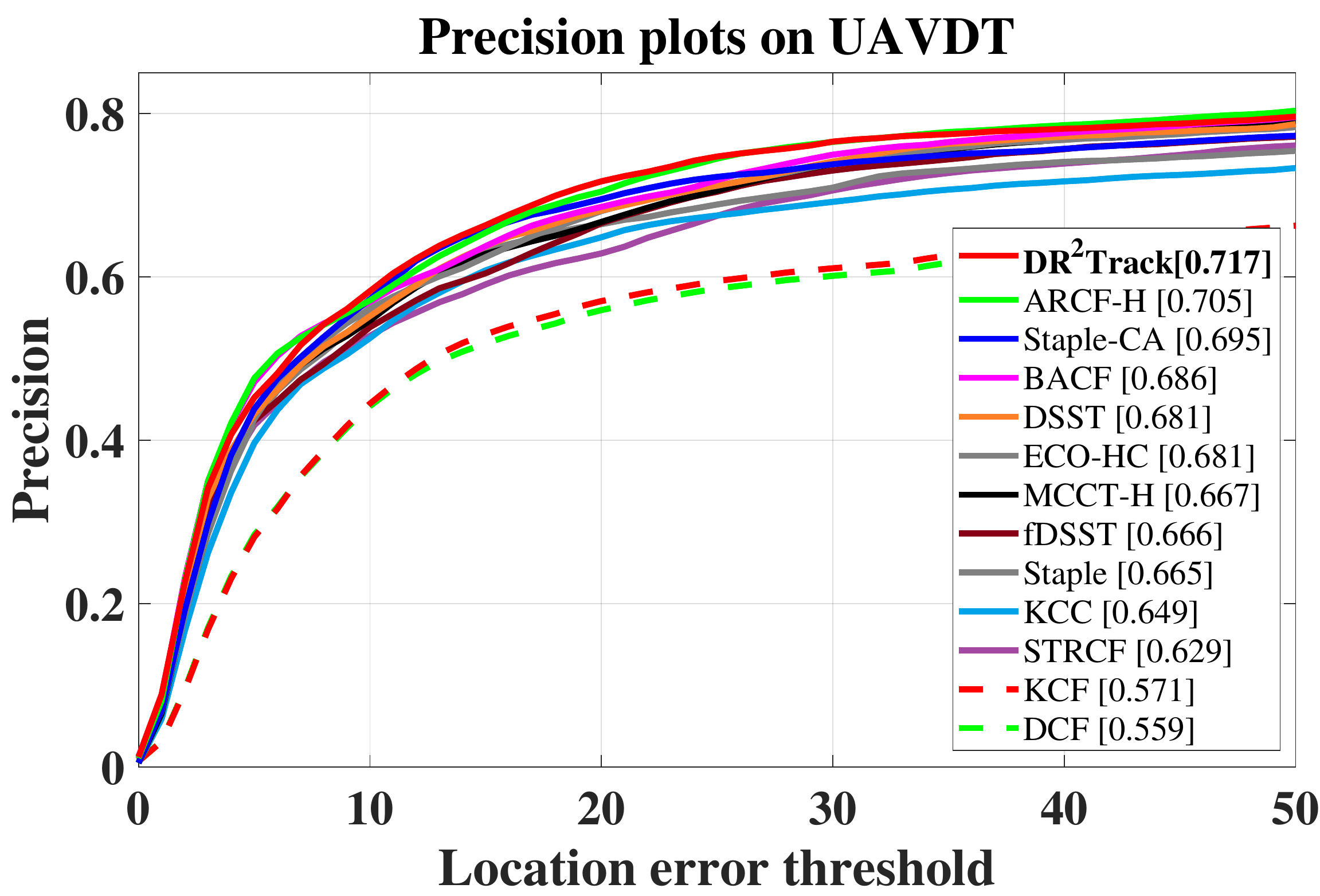}
				\\
				\includegraphics[width=1\columnwidth]{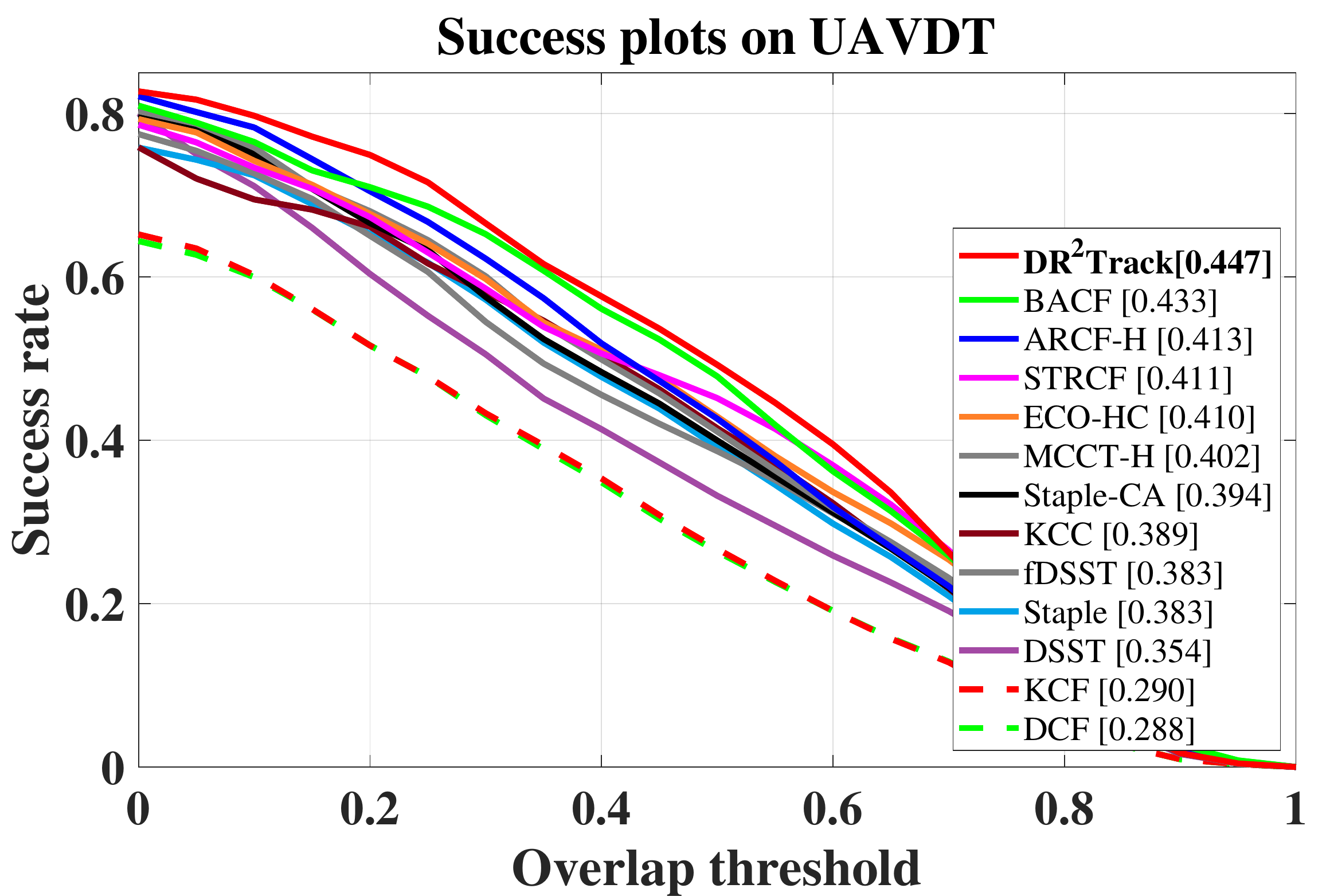}
			\end{minipage}
		}
	\end{center}
	\caption{Overall performance of hand-crafted real-time trackers on  UAV123@10fps~\cite{Mueller2016ECCV}, DTB70~\cite{Li2017AAAI}, and UAVDT~\cite{Du2018ECCV}. Precision plots and success plots can respectively exhibit the percentage of all evaluated frames in which the distance of estimated location with ground truth one is smaller than distinctive thresholds, and in which the overlap between predicted bounding box and ground truth one is greater than distinctive thresholds. The score at 20 pixel and area under curve (AUC) are respectively employed for ranking. DR$^2$Track has the best performance on precision and success rate in all benchmarks.}
	\label{fig:overall}
\end{figure*}
\begin{table*}[!h]
	\setlength{\tabcolsep}{0.4mm}
	\centering
	\caption{Average precision, AUC and speed comparison between top 12 hand-crafted trackers on UAV123@10fps \cite{Mueller2016ECCV}, UAVDT \cite{Du2018ECCV} and DTB70 \cite{Li2017AAAI}. \textcolor[rgb]{ 1,  0,  0}{Red}, 
		\textcolor[rgb]{ 0,  1,  0}{green} and \textcolor[rgb]{ 0,  0,  1}{blue} color indicate the first, second and third place, respectively.}
	\vspace{-0.3cm}
	\begin{tabular}{cc c c c||cc c c c||cc c c c}
		\hline\hline
		Tracker&Precision&AUC&Speed&Venue&Tracker&Precision&AUC&Speed&Venue&Tracker&Precision&AUC&Speed&Venue\\ 		
		\hline
		\textbf{DR$^2$Track}&\textcolor[rgb]{ 1,  0,  0}{\textbf{0.710}}&\textcolor[rgb]{ 1, 0,0}{\textbf{0.474}}&46.1&Ours&STRCF\cite{li2018learning}&0.635&0.435&28.5&CVPR'18&BACF\cite{galoogahi2017learning}&0.616&0.416&56.0&CVPR'17\\
		ARCF-H\cite{huang2019learning}&{0.641}&{{0.421}}&51.2&ICCV'19&ECO-HC\cite{danelljan2017eco}&0.653&\textcolor[rgb]{ 0,  0,  1}{\textbf{0.442}}&\textcolor[rgb]{ 0,  1,  0}{\textbf{69.3}}&CVPR'17&fDSST\cite{danelljan2017discriminative}&0.572&0.373&\textcolor[rgb]{ 1,  0,  0}{\textbf{168.0}}&PAMI'17\\
		SRDCF\cite{danelljan2015learning}&0.582&0.402&14.0&ICCV'15&ARCF-HC\cite{huang2019learning}&\textcolor[rgb]{ 0,  1,  0}{\textbf{0.693}}&\textcolor[rgb]{ 0,  1,  0}{\textbf{0.468}}&15.3&ICCV'19&MCCT-H\cite{wang2018multi}&0.622&0.413&\textcolor[rgb]{ 0,  0,  1}{\textbf{59.7}}&CVPR'18 \\
		CSR-DCF\cite{lukezic2017discriminative}&\textcolor[rgb]{ 0,  0,  1}{\textbf{0.654}}&0.426&12.1&CVPR'17&SRDCFdecon\cite{danelljan2016adaptive}&0.577&0.397&7.5&CVPR'16&STAPLE-CA\cite{mueller2017context}&0.595&0.388&58.9&CVPR'17\\
		\hline\hline
	\end{tabular}%
	\label{tab:2}
\end{table*}%
including over 90K frames recorded in all kinds of difficult aerial scenarios. For comparison, we employ sixteen state-of-the-art (SOTA) hand-crafted trackers, \emph{i.e.}, KCC~\cite{wang2018kernel}, DCF~\cite{henriques2015high}, KCF~\cite{henriques2015high}, DSST~\cite{danelljan2017discriminative}, fDSST~\cite{danelljan2017discriminative}, STAPLE~\cite{bertinetto2016staple}, MCCT-H\cite{wang2018multi}, ECO-HC\cite{danelljan2017eco}, BACF\cite{galoogahi2017learning}, STAPLE-CA\cite{mueller2017context}, STRCF\cite{li2018learning}, ARCF-H\cite{huang2019learning}, SRDCF\cite{danelljan2015learning}, SRDCFdecon\cite{danelljan2016adaptive}, CSR-DCF\cite{lukezic2017discriminative}, ARCF-HC\cite{huang2019learning}, and eleven deep-based trackers, \emph{i.e.}, ECO \cite{danelljan2017eco}, C-COT \cite{danelljan2016beyond}, MCCT \cite{wang2018multi}, DeepSTRCF \cite{li2018learning}, IBCCF \cite{li2017integrating}, ASRCF \cite{dai2019visual}, SiameseFC \cite{bertinetto2016fully}, CFNet \cite{valmadre2017end}, TADT \cite{li2019target}, UDT+ \cite{wang2019unsupervised}, DSiam \cite{guo2017learning}. Notice that the  same platform stated in \ref{details} is used for all experiments in this work and the codes released by the authors are adopted without any modification.

\begin{subsection}{Implementation details}\label{details}
	\textbf{Platform.}
	The experiments are executed using MATLAB R2018a on a computer with an i7-8700K CPU (3.7GHz), 32GB RAM and an NVIDIA GTX 2080 GPU.  
	
	\textbf{Parameters.} For the hyper parameters of ADMM, we set $\gamma_{max}=10000, \beta=10, \gamma^0=1$, and the number of iteration $E$ is set as 4. The number of top local maximums $N$ is 30 and $\mu$ is set as 0.25. $\theta$ is set to be 12. The search region of the proposed tracker is set to be a square with a size of $5\sqrt{WH}$, where $H$ and $W$ denote the height and width of the object. Note that all the parameters of the presented tracker are left unchanged in all experiments for a fair comparison.

	\textbf{Features.} Only the hand-crafted features, \emph{i.e.}, gray-scale, histogram of oriented gradient (HOG) \cite{Dalal2005CVPR} and color names (CN) \cite{danelljan2014adaptive}, are utilized for appearance representations. The cell size for feature extraction is set to be $4\times 4$. 
	
	\textbf{Scale.} For the scale estimation, the scale filter \cite{danelljan2017discriminative} trained on a multi-scale pyramid of the object is applied to select the optimal scale. 
\end{subsection}

\begin{subsection}{Comparison with hand-crafted trackers}
	Using hand-crafted features, hand-crafted trackers have calculation feasibility on a single CPU. Among them, real-time trackers run at over 30 frames per second (fps) that is generally the capture frequency of drone cameras.
\begin{subsubsection}{Overall performance evaluation}
 	Adopting one-pass evaluation protocol~\cite{wu2013online}, two quantitative measures, \emph{i.e.}, precision, and success rate, are used to evaluate tracking performance. The comparison results between DR$^2$Track and other real-time trackers on three challenging UAV benchmarks are displayed in Fig.~\ref{fig:overall}. Clearly, DR$^2$Track outperforms other real-time hand-crafted trackers in precision and success rate on three benchmarks. In UAV123@10fps \cite{Mueller2016ECCV}, DR$^2$Track has improved the tracking precision by a large margin, \emph{i.e.}, over 10\% than the second-best ECO-HC\cite{danelljan2017eco}, which utilizes many techniques, such as efficient convolution operator, sample generative model, and sparser updating scheme. In DTB70 \cite{Li2017AAAI}, compared with our baseline tracker STRCF \cite{li2018learning} (second-best in precision and third place in AUC), DR$^2$ Track has a notable improvement of 9.2\% and 10.5\%, respectively in precision and success rate. In UAVDT \cite{Du2018ECCV}, DR$^2$Track exhibits stably excellent performance in terms of both precision and AUC.  Some qualitative tracking results are shown in Fig.~\ref{fig:display}.
 	The average tracking performances on three benchmarks, \emph{i.e.}, precision, AUC, and speed, of the top 12 hand-crafted trackers are listed in Table~\ref{tab:2}. 
 	Compared with other SOTA hand-crafted trackers, the presented tracker in this work ranks first on the average precision and AUC. Compared with the second-best hand-crafted tracker ARCF-HC, DR$^2$Track has obtained an improvement of 2.5\% and 201.3\% in average precision and speed. 
 Based on the STRCF tracker~\cite{li2018learning}, an average improvement of 11.8\% in precision and 8.9\% in AUC has been achieved by DR$^2$Track, while the tracking speed has been increased by 61.7\%, proving DR$^2$Track's extraordinary tracking performance and efficiency.

	\noindent\textbf{\emph{Remark 2}:} Due to larger time gap between successive frames, the selected UAV123@10fps dataset \cite{Mueller2016ECCV} from the recorded 30fps one has more fast motion situations, and DTB70 \cite{Li2017AAAI} also covers various drone motion cases. In these two benchmarks, the prominent improvement of DR$^2$Track compared to the others has validated its robustness to UAV/object motion.
\end{subsubsection} 
	\begin{figure*}[!t]
	\begin{center}

		\subfigure { \label{fig:att1} 
			\begin{minipage}{0.23\textwidth}
				\centering
				\includegraphics[width=1\columnwidth]{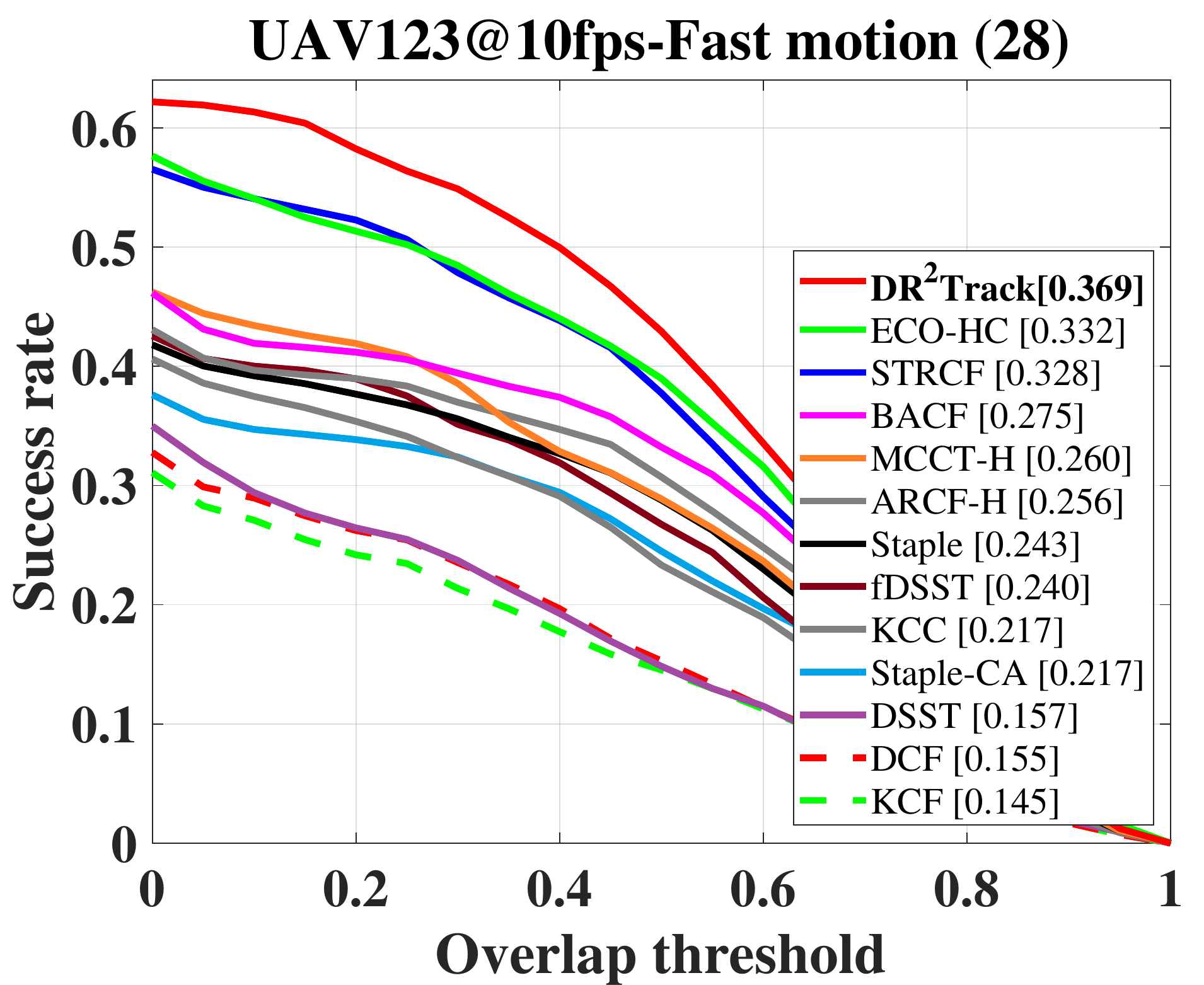}
			\end{minipage}
		}
		\subfigure { \label{fig:att2} 
			\begin{minipage}{0.23\textwidth}
				\centering
				\includegraphics[width=1\columnwidth]{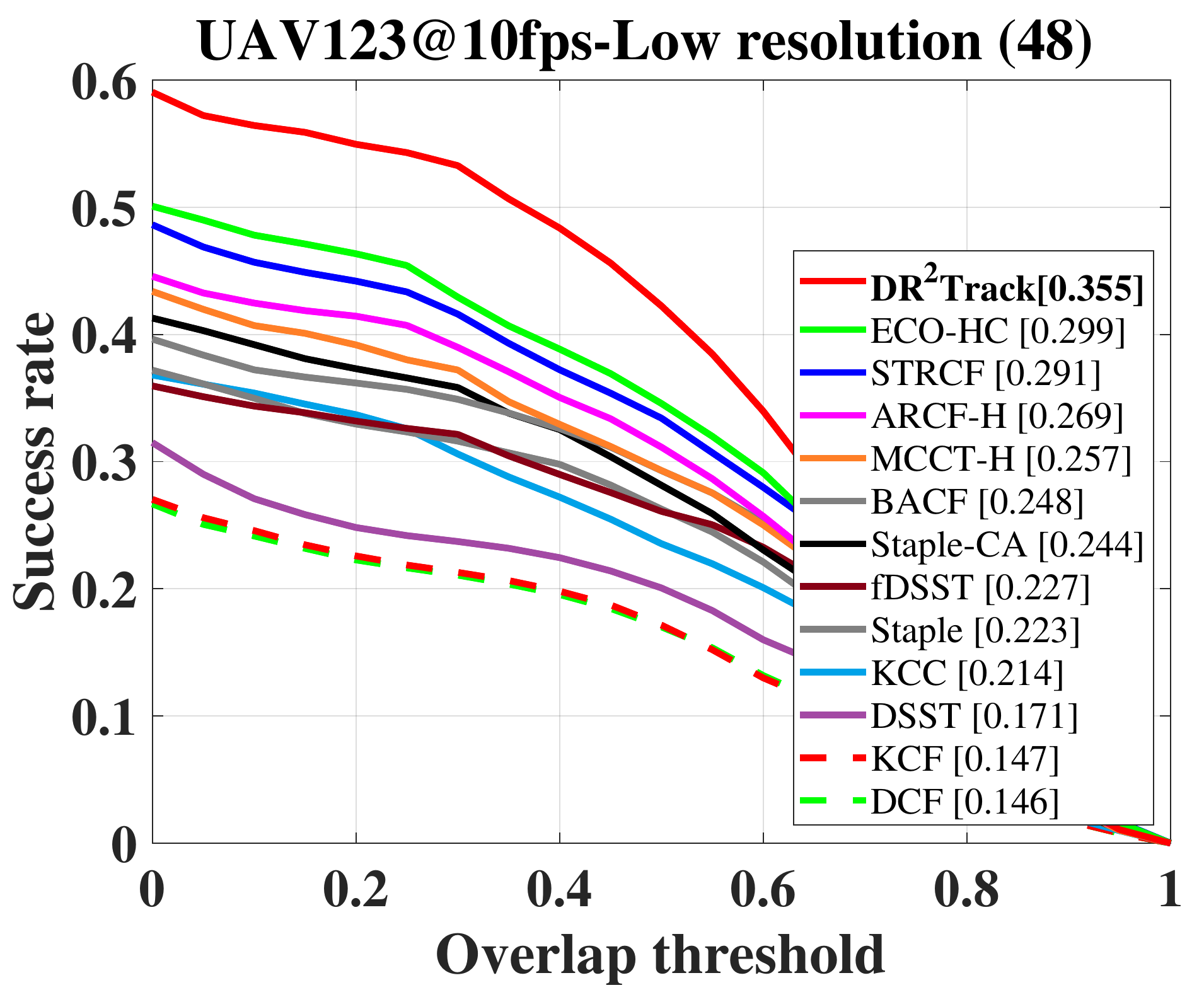}
			\end{minipage}
		}
		\subfigure { \label{fig:att3} 
			\begin{minipage}{0.23\textwidth}
				\centering
				\includegraphics[width=1\columnwidth]{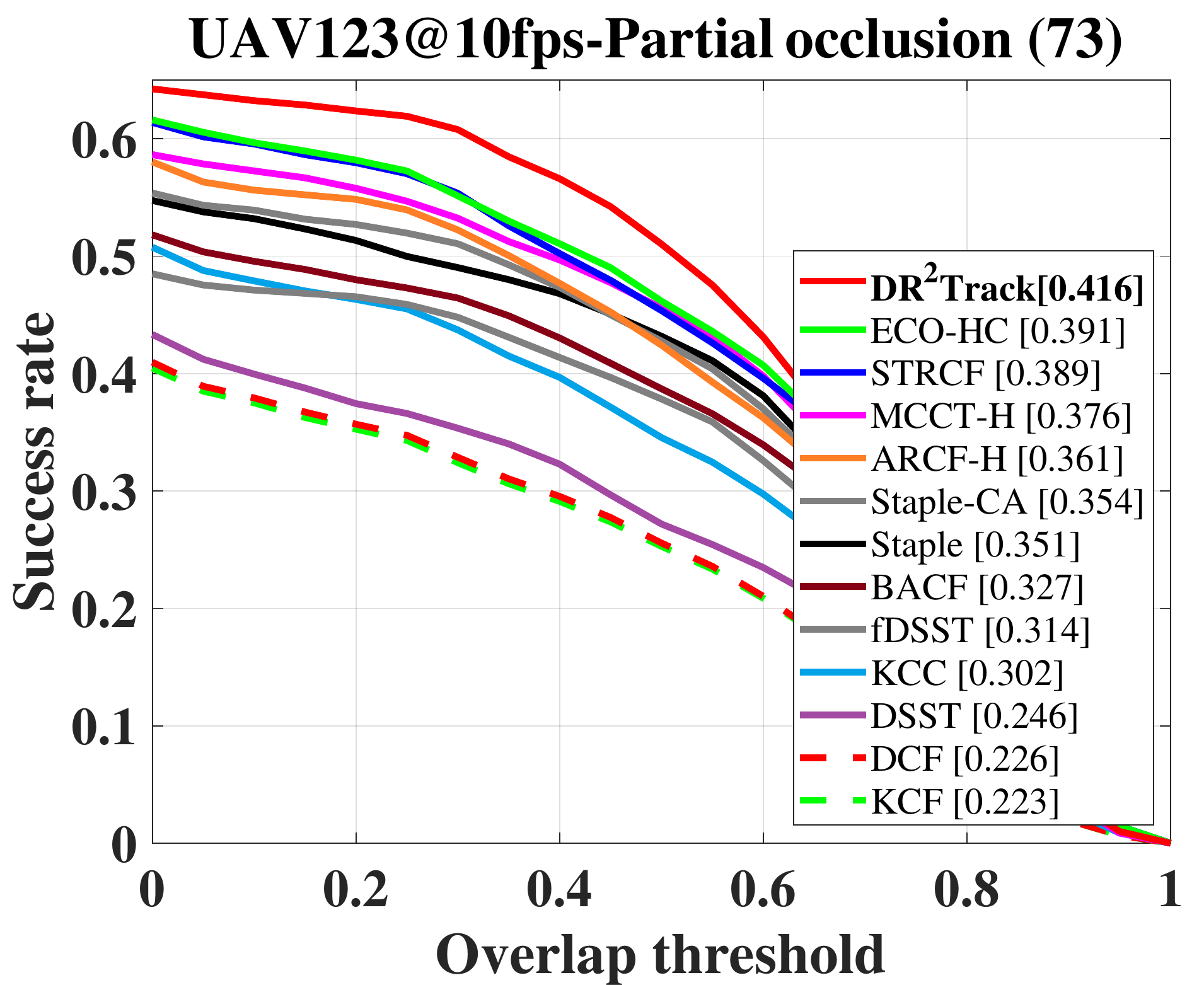}
			\end{minipage}
		}
		\subfigure { \label{fig:att4} 
			\begin{minipage}{0.23\textwidth}
				\centering
				\includegraphics[width=1\columnwidth]{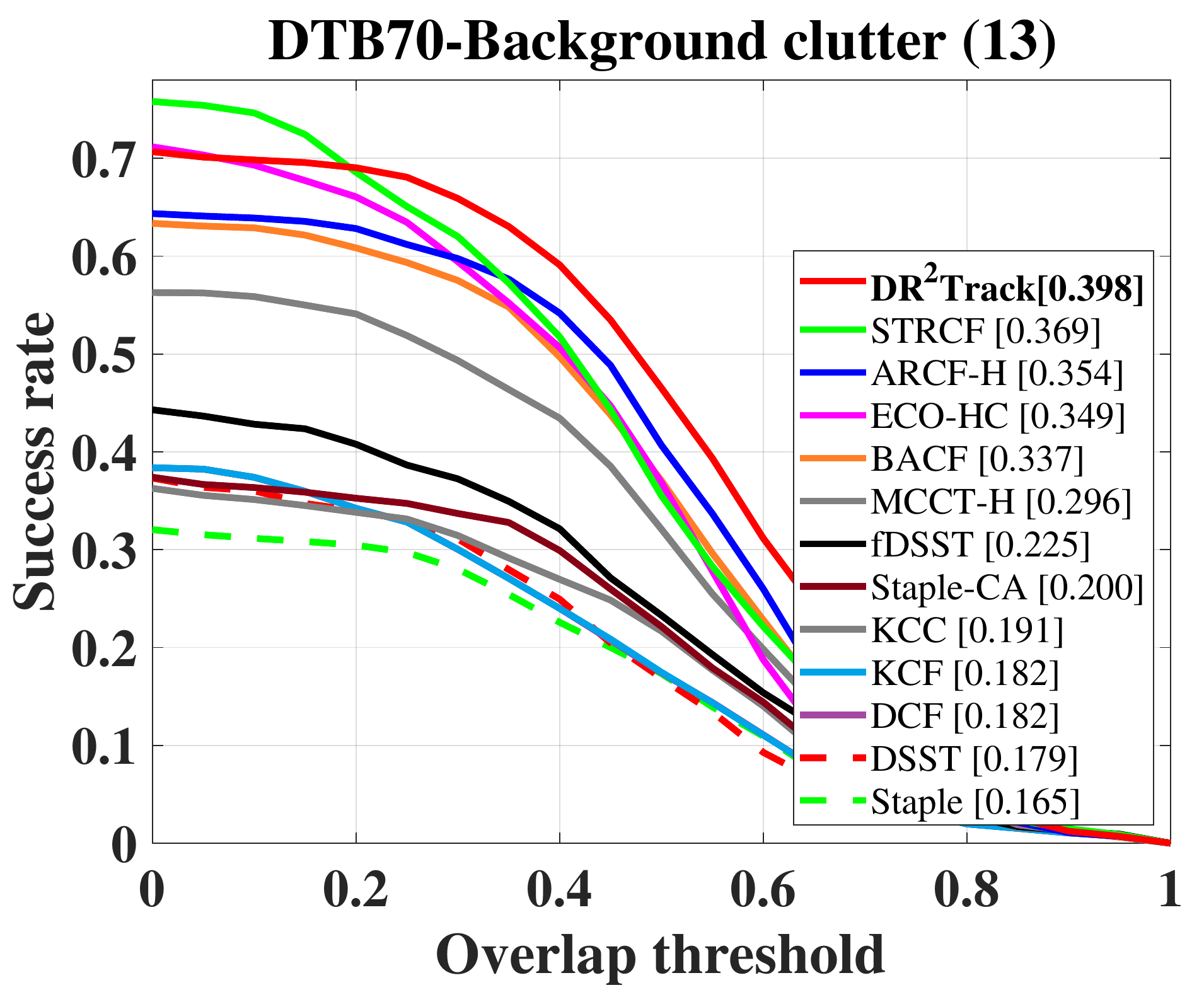}
			\end{minipage}
		}
		\subfigure { \label{fig:att5} 
			\begin{minipage}{0.23\textwidth}
				\centering
				\includegraphics[width=1\columnwidth]{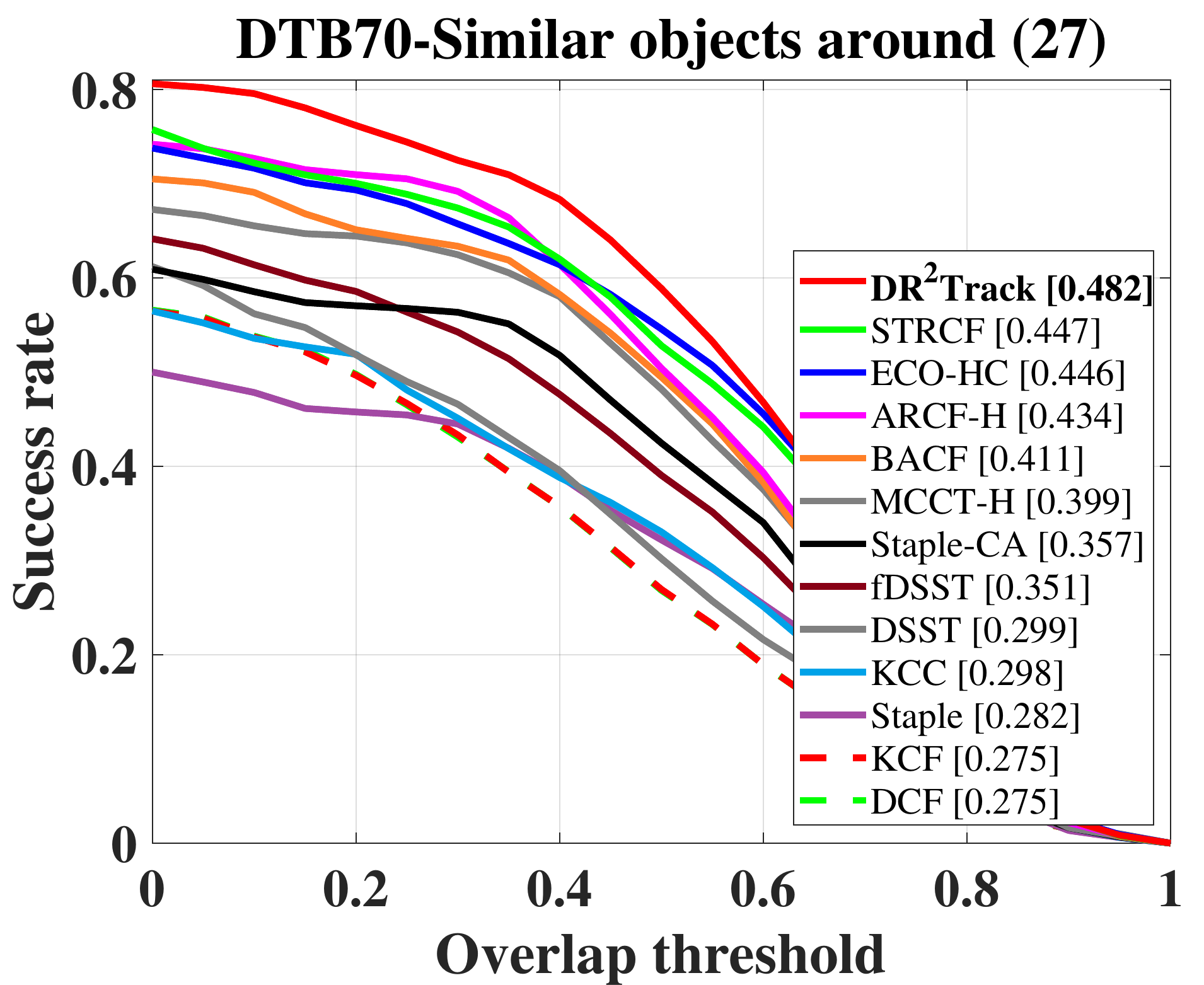}
			\end{minipage}
		}
		\subfigure { \label{fig:att6} 
			\begin{minipage}{0.23\textwidth}
				\centering
				\includegraphics[width=1\columnwidth]{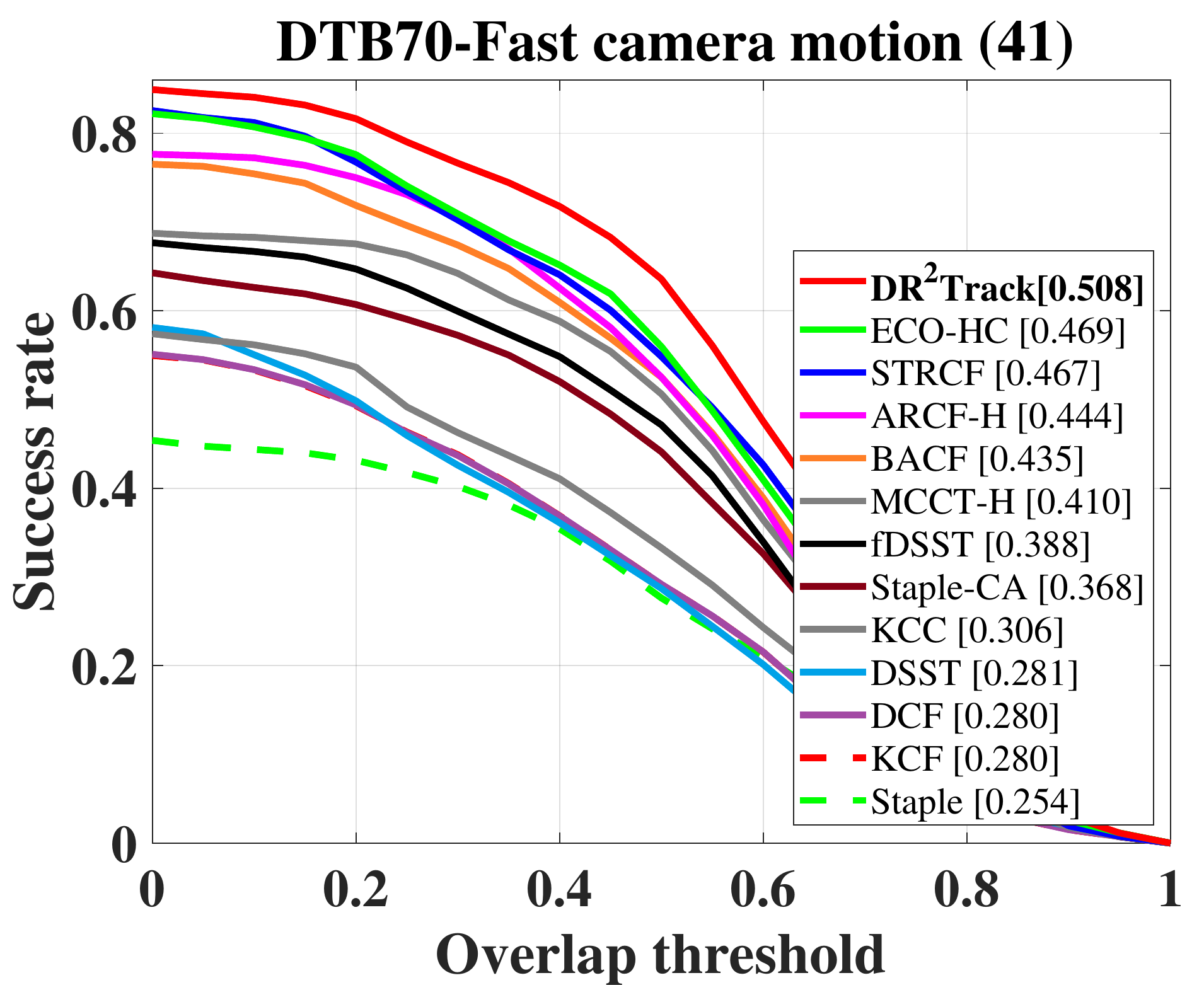}
			\end{minipage}
		}
		\subfigure { \label{fig:att7} 
			\begin{minipage}{0.23\textwidth}
				\centering
				\includegraphics[width=1\columnwidth]{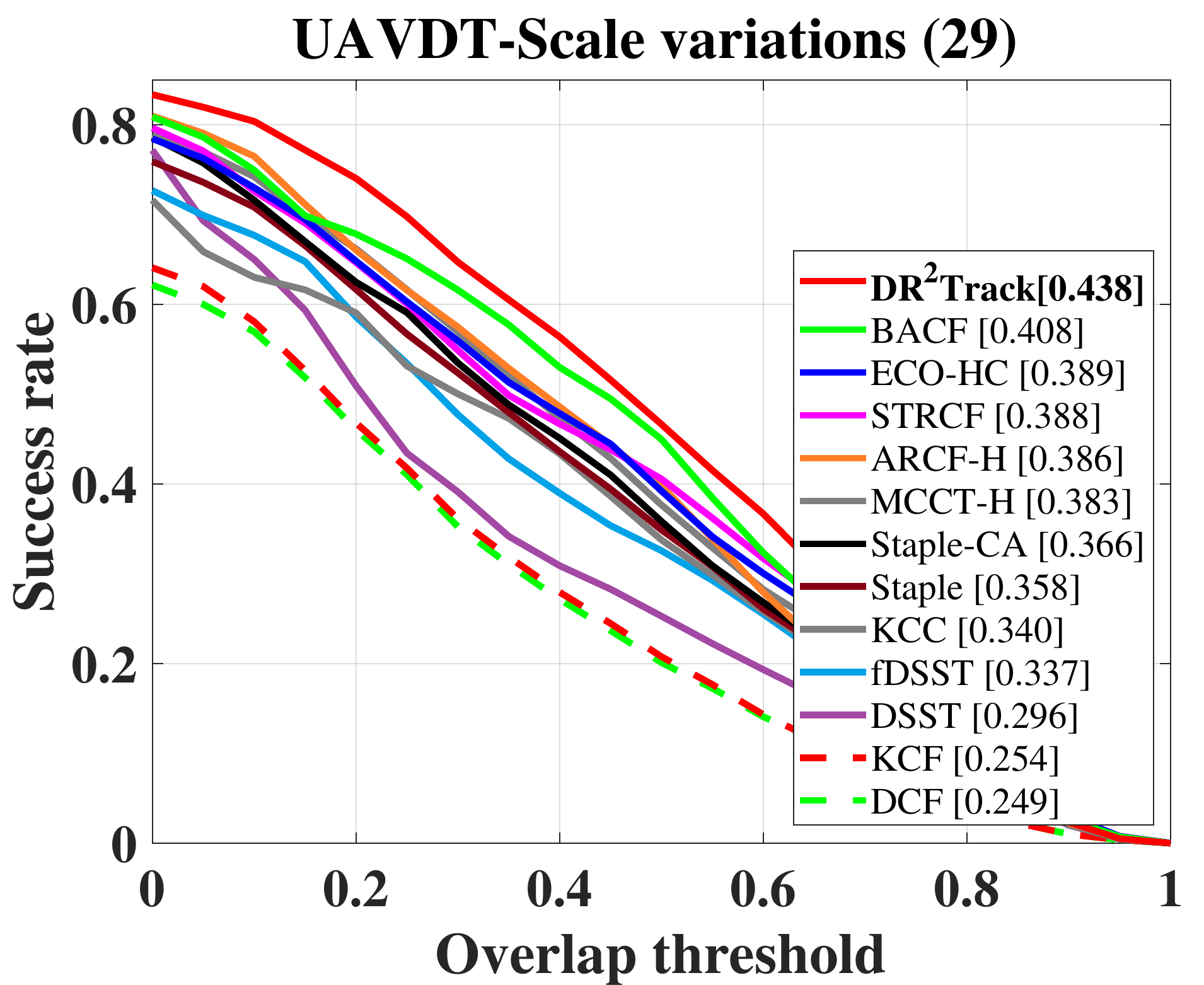}
			\end{minipage}
		}
		\subfigure { \label{fig:att8} 
			\begin{minipage}{0.23\textwidth}
				\centering
				\includegraphics[width=1\columnwidth]{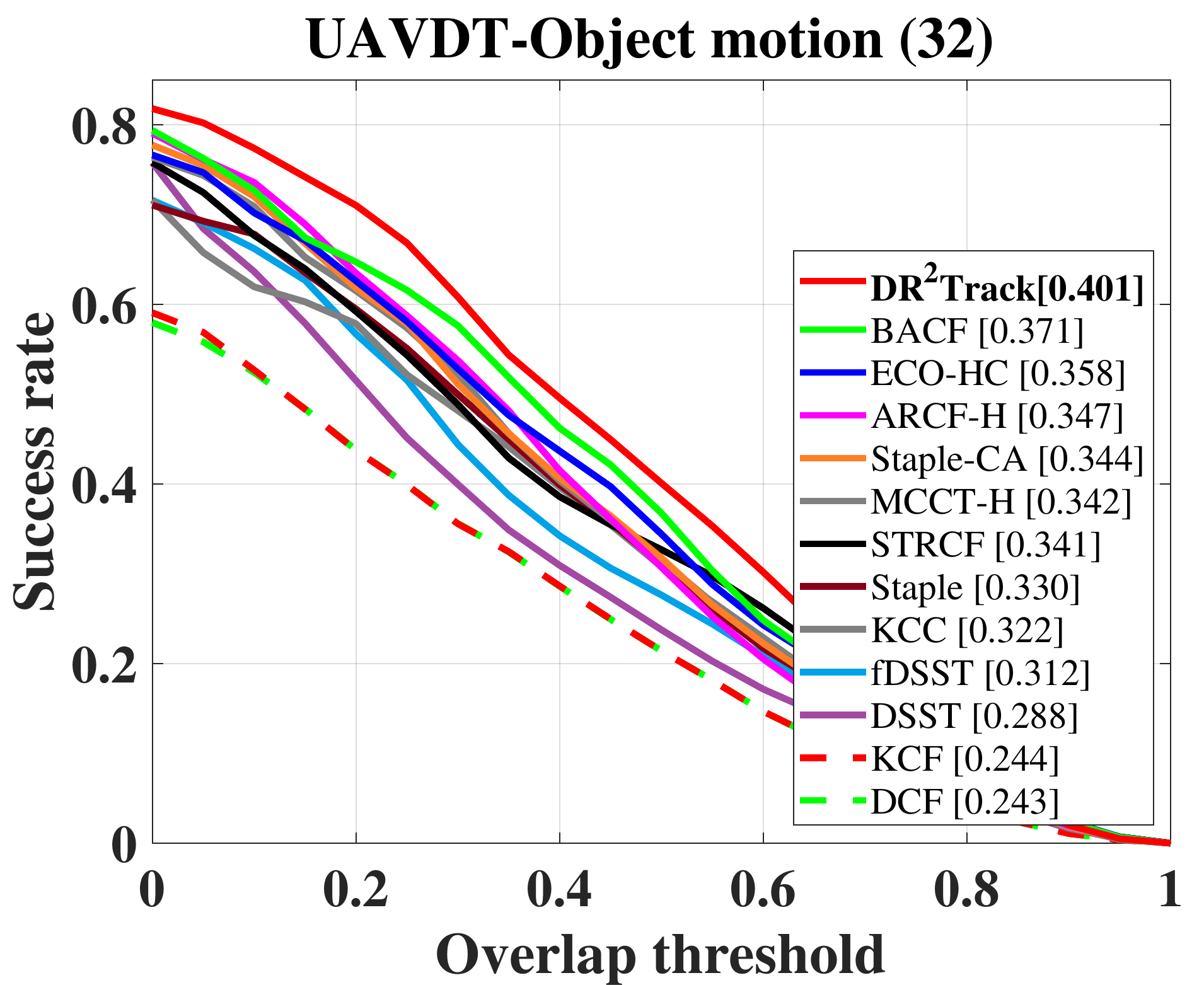}
			\end{minipage}
		}
	\end{center}
	
	\caption{Attribute-based comparison with hand-crafted real-time trackers. Eight attributes, \emph{i.e.}, fast motion, low resolution, partial occlusion, background clutter, similar objects around, fast camera motion, scale variations and object motion from three UAV datasets are displayed.}
	\label{fig:attribute}
\end{figure*}
\begin{subsubsection}{Attribute-based comparison}
Attribute-based comparison is conducted on eight challenging scenarios. Their success rate plots are displayed in Fig.~\ref{fig:attribute}. In situations of background clutter and similar objects around, DR$^2$Track has improved STRCF by 7.8\% in terms of AUC. This improvement stems from the effective distractor repression which can automatically detect the surrounding objects with similar appearance, and repress them once they emerge. In fast motion, fast camera motion and object motion cases, DR$^2$Track has respectively obtained an advantage of 11.1\%, 8.3\%, and 8.0\% compared with the second-place trackers, proving its robustness against motion owing to the search area enlargement as well as prediction. In other attributes, \emph{i.e.}, low resolution, partial occlusion, and scale variations, DR$^2$Track has achieved a remarkable advantage of 18.7\%, 6.3\%, and 7.3\%, validating its superior generalization ability.

\end{subsubsection}

\end{subsection}
	\begin{figure}[!b]
	\begin{center}

		\subfigure{\label{fig:deep1}
			\begin{minipage}{0.225\textwidth}
				\centering
				\includegraphics[width=1\columnwidth]{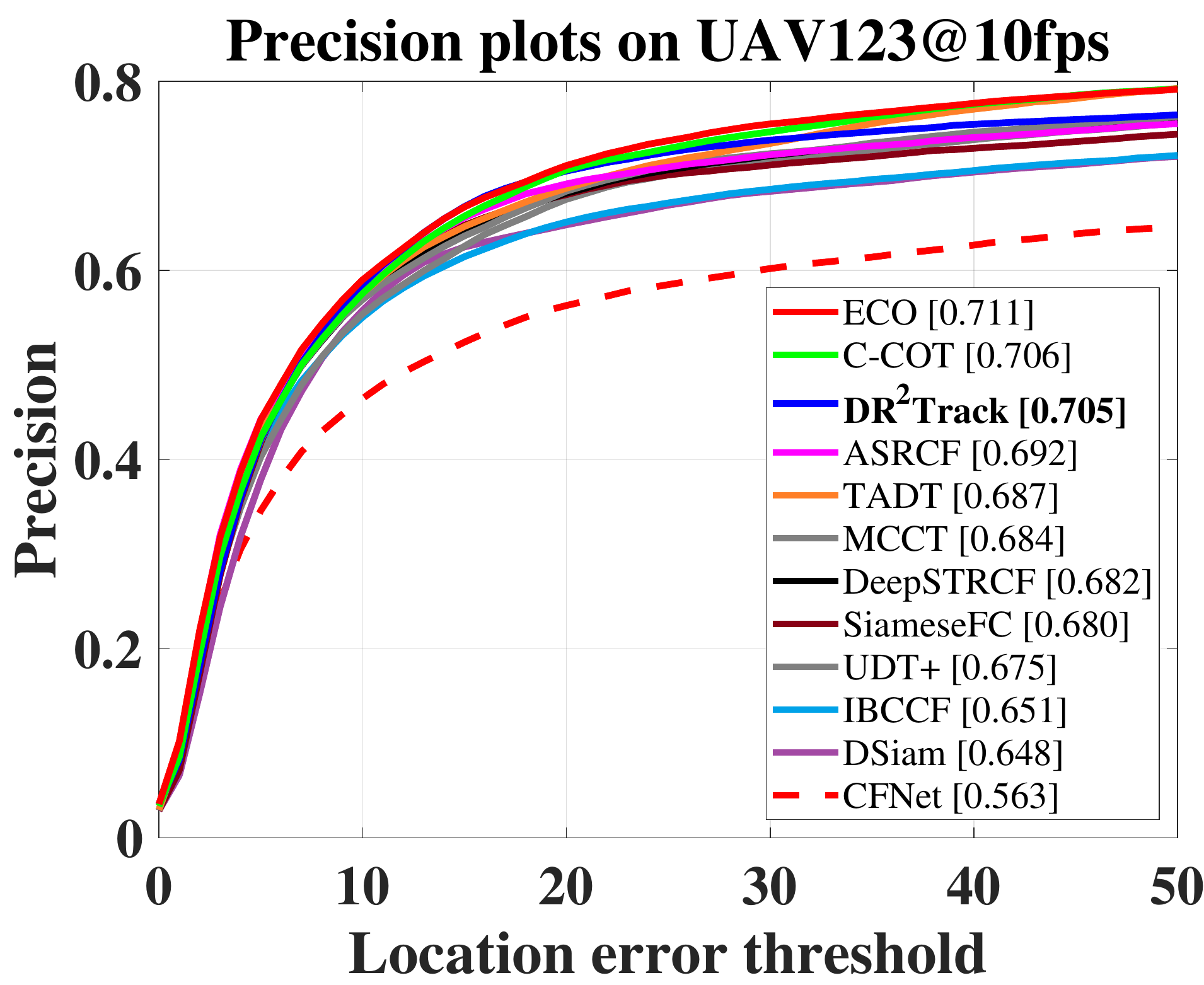}
			\end{minipage}
		}
		\subfigure{ \label{fig:deep2} 
			\begin{minipage}{0.225\textwidth}
				\centering
				\includegraphics[width=1\columnwidth]{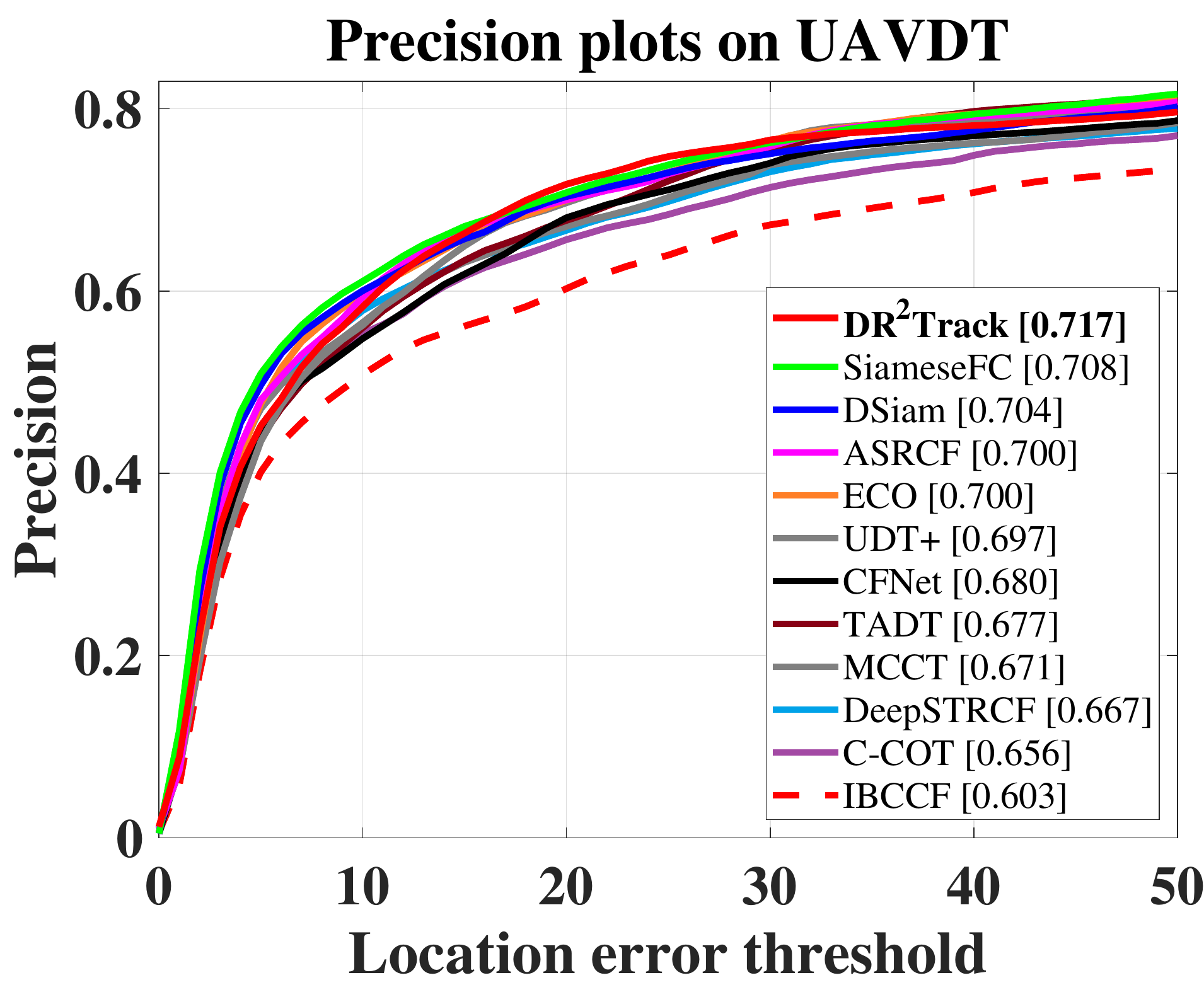}
			\end{minipage}
		}
	\end{center}
	
	\caption{Precision plots on UAV123@10fps and UAVDT of the present approach compared with eleven deep-based trackers.}
	\label{fig:deep}
\end{figure}
\begin{subsection}{Comparison with deep trackers}
	This subsection compared DR$^2$Track with deep trackers, including six DCF type trackers using deep features, \emph{i.e.}, ECO \cite{danelljan2017eco}, C-COT \cite{danelljan2016beyond}, ASRCF \cite{dai2019visual}, MCCT \cite{wang2018multi}, DeepSTRCF \cite{li2018learning}, IBCCF \cite{li2017integrating} and five DNN architecture based trackers: SiameseFC \cite{bertinetto2016fully}, TADT \cite{li2019target}, UDT+ \cite{wang2019unsupervised}, CFNet \cite{valmadre2017end}, and DSiam \cite{guo2017learning}. Only with hand-crafted features, DR$^2$Track obtained the third place on precision in UAV123@10fps \cite{Mueller2016ECCV} and the first place in UAVDT \cite{Du2018ECCV}, as shown in Fig.~\ref{fig:deep}. The comparison of average precision and speed are reported in Table~\ref{tab:1}:  DR$^2$Track ranks first in precision. Implemented on a cheap CPU, DR$^2$Track is faster than most deep trackers running on a high-end GPU and improves the speed by 220\% compared to the second-place ECO\cite{danelljan2017eco}.

	\end{subsection}
\begin{subsection}{Ablation Studies}
	\begin{subsubsection}{Analysis of separate components}
		 To prove the effectiveness of each component proposed in our work, we compare DR$^2$Track to its different versions with distinctive components added. It is noted that '-MA' denotes the method only applying motion-aware (MA) search strategy, while '-DR' means using dynamic regression for the filter training. The evaluation results are displayed in Table~\ref{tab:3}, from which it can be clearly seen that the proposed two components have boosted the tracking performance by a large margin.  
	\end{subsubsection}

\begin{figure}[!b]
	\begin{center}

		\subfigure { \label{fig:hyper1} 
			\begin{minipage}{0.49\textwidth}
				\centering
				\includegraphics[width=1\columnwidth]{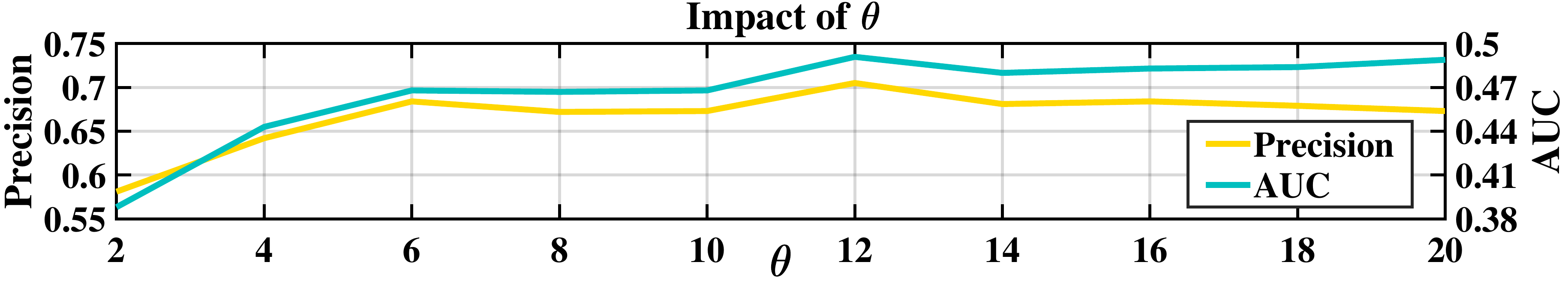}
			\end{minipage}
		}
		\subfigure { \label{fig:hyper2} 
			\begin{minipage}{0.49\textwidth}
				\centering
				\includegraphics[width=1\columnwidth]{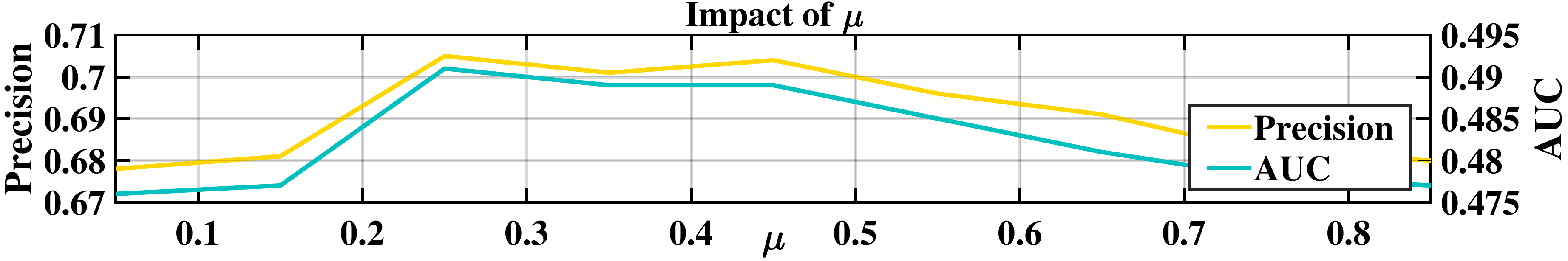}
			\end{minipage}
		}
		\subfigure { \label{fig:hyper3} 
			\begin{minipage}{0.49\textwidth}
				\centering
				\includegraphics[width=1\columnwidth]{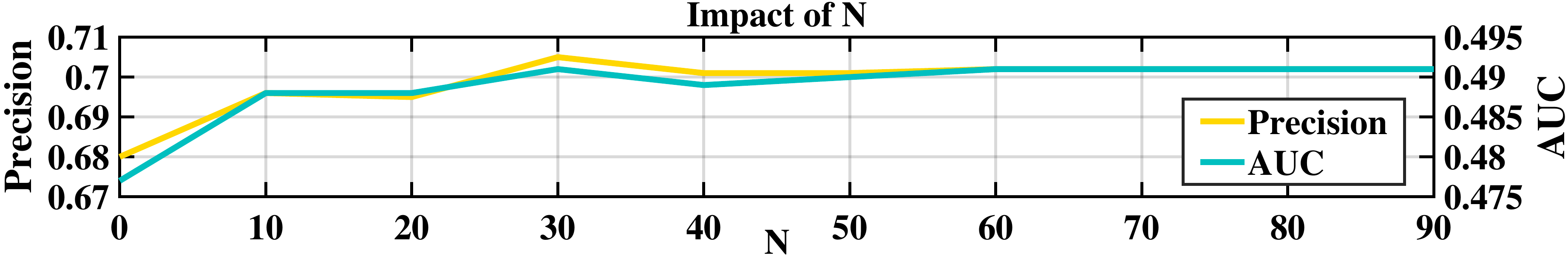}
			\end{minipage}
		}
	\end{center}
	\caption{Sensitivity analysis of three parameters ($\theta$, $\mu$,
		and $N$) on UAV123@10fps\cite{Mueller2016ECCV}. The variations of $N$ and $\mu$ have a relatively small impact on tracking performance (the precision and AUC are mostly within the range of 0.67 to 0.71 and 0.475 to 0.495, respectively.), while the change of $\theta$ has a larger influence.}
	\label{fig:hyper}
\end{figure}

\begin{figure*}[!t]
	\centering
	\includegraphics[width=1.97\columnwidth]{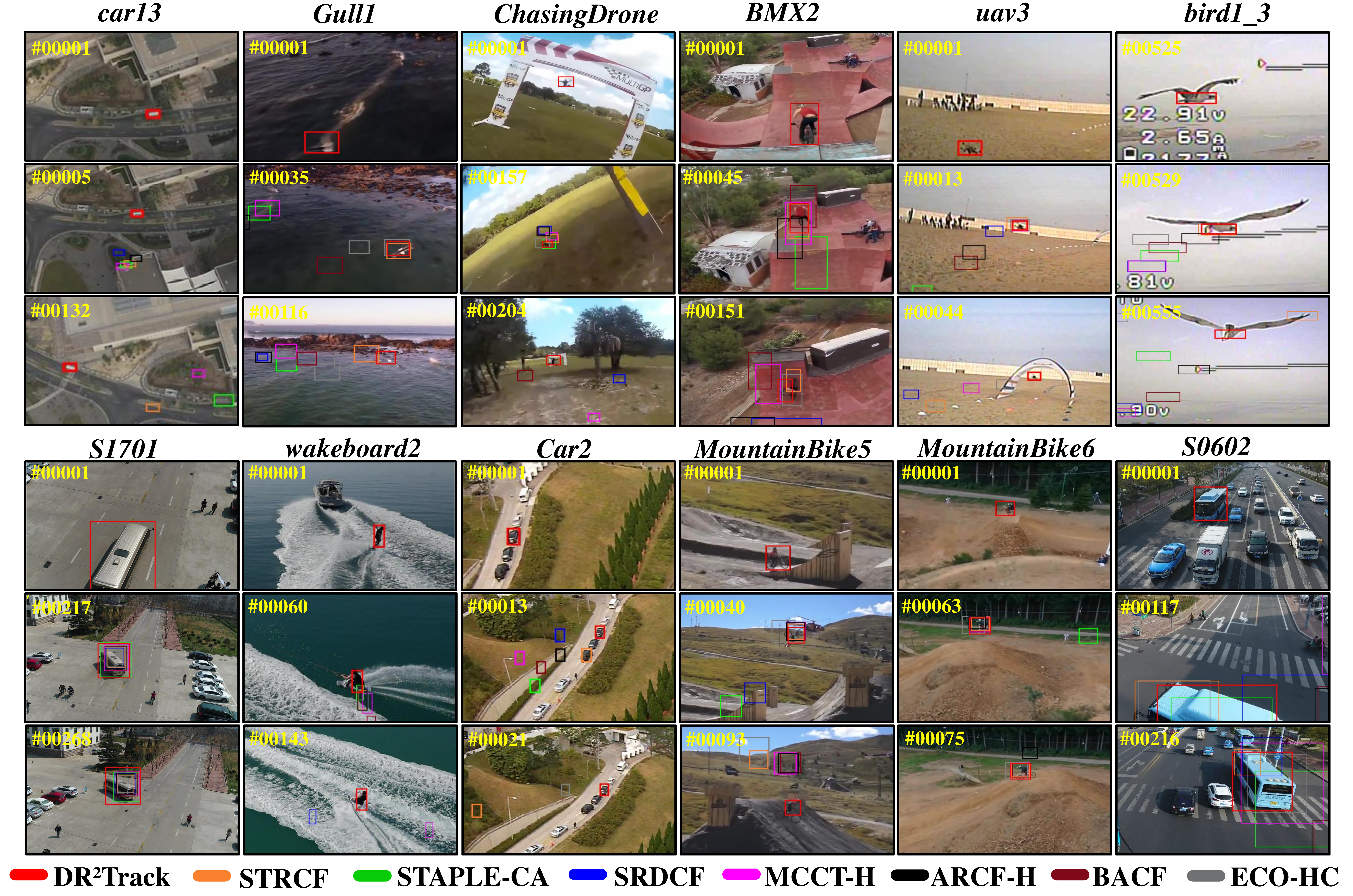}
	\caption{Tracking results demonstration of eight hand-crafted  trackers on twelve video sequences, \emph{i.e.}, \emph{car13}, \emph{Gull1}, \emph{ChasingDrones}, \emph{BMX2}, \emph{uav3}, \emph{bird1\_3}, \emph{S1701}, \emph{wakeboard2}, \emph{Car2}, \emph{MountainBike5}, \emph{MountainBike6} and \emph{S0602}. Video and code can be found on \url{https://youtu.be/HQz_xIu9XOc} and \url{http://github.com/vision4robotics/DR2Track}, respectively. }
	\label{fig:display}
\end{figure*}
	\begin{table*}[!t]
	\setlength{\tabcolsep}{1mm}
	\centering
	\caption{Precision and speed comparison between DR$^2$Track on UAVDT\cite{Du2018ECCV} and UAV123@10fps\cite{Mueller2016ECCV} with deep trackers. * indicates GPU speed.  \textcolor[rgb]{ 1,  0,  0}{Red}, \textcolor[rgb]{ 0,  1,  0}{green} and \textcolor[rgb]{ 0,  0,  1}{blue} respectively represent the first, second and third place.}
	\vspace{-0.3cm}
	\begin{tabular}{cccccccccccccccc}
		\hline\hline
		Tracker&\textbf{DR$^2$Track}&TADT&C-COT&UDT+&SiameseFC&IBCCF&CFNet&ECO&DSiam&MCCT&DeepSTRCF&ASRCF\\
		\hline
		Precision&\textcolor[rgb]{ 1,  0,  0}{\textbf{0.711}}&0.682&0.681&0.675&0.681&0.627&0.622&\textcolor[rgb]{ 0, 1,  0}{\textbf{0.706}}&0.676&0.678&0.675&\textcolor[rgb]{ 0,  0,  1}{\textbf{0.696}}\\
		Speed&\textcolor[rgb]{ 0,  1,  0}{\textbf{46.5}}&34.7*&1.1*&\textcolor[rgb]{ 1,  0,  0}{\textbf{49.8}}*&37.9*&2.9*&\textcolor[rgb]{ 0,  0,  1}{\textbf{41.1}}*&14.5*&15.9*&8.6*&6.4*&24.1	\\	Venue&Ours&CVPR'19&ECCV'16&CVPR'19&ECCV'16&CVPR'17&CVPR'17&CVPR'17&CVPR'17&CVPR'18&CVPR'18&CVPR'19\\
		\hline\hline
	\end{tabular}%
	\label{tab:1}
\end{table*}%

\begin{subsubsection}{Impacts of hyper parameters}
		Three hyper parameters, \emph{i.e.}, temporal regularization parameter $\theta$, distractors repression factor $\mu$, and the number of local maximums for suppression $N$, are investigated for sensitivity analysis as follows. How the different values of hyper parameters affect the tracking performance of DR$^2$Track can be seen in Fig.~\ref{fig:hyper}. It is noted that we fix the other two hyper parameters when changing the value of the analyzed one. In terms of $\theta$, the tracking performance drops fast as the value descends since decreasing temporal constraint can introduce noise easily. 

	\end{subsubsection}
\end{subsection}
\begin{table}[!t]
	\setlength{\tabcolsep}{1mm}
	\centering
	\caption{Precision and AUC comparison between trackers with different components enabled on UAV123@10fps benchmark}
	\vspace{-0.2cm}
	\begin{tabular}{ccccc}
		\hline\hline
		Tracker&DR$^2$Track&DR$^2$Track-DR&DR$^2$Track-MA&STRCF\\
		\hline
		Precision&\textbf{0.705}&0.691&0.680&0.635\\
		AUC&\textbf{0.491}&0.481&0.477&0.435\\
		\hline\hline
	\end{tabular}%
	\label{tab:3}
\end{table}%
\section{CONCLUSIONS}\label{sec:CONCLUSIONS}
Different from the classical DCF trackers with a fixed training label, the regression target response in this work can be dynamically changed to upgrade the training quality. By introducing local maximums of response maps, distractors can be located and repressed with pixel-level penalization. Extensive experiment results on three difficult UAV benchmarks show that  DR$^2$Track has outstanding performance with real-time frame rates on a cheap CPU. We strongly believe that our work can contribute to the prosperity of visual object tracking and its applications in the field of UAV.

\section*{ACKNOWLEDGMENT}
The work is supported by the National Natural Science Foundation of China under Grant 61806148 and the State Key Laboratory of Mechanical Transmissions (Chongqing
University) under Grant SKLMT-KFKT-201802.


\bibliographystyle{IEEEtran}  
\bibliography{IEEEabrv,ref}

\begin{thebibliography}{10}
\providecommand{\url}[1]{#1}
\csname url@rmstyle\endcsname
\providecommand{\newblock}{\relax}
\providecommand{\bibinfo}[2]{#2}
\providecommand\BIBentrySTDinterwordspacing{\spaceskip=0pt\relax}
\providecommand\BIBentryALTinterwordstretchfactor{4}
\providecommand\BIBentryALTinterwordspacing{\spaceskip=\fontdimen2\font plus
\BIBentryALTinterwordstretchfactor\fontdimen3\font minus
  \fontdimen4\font\relax}
\providecommand\BIBforeignlanguage[2]{{%
\expandafter\ifx\csname l@#1\endcsname\relax
\typeout{** WARNING: IEEEtran.bst: No hyphenation pattern has been}%
\typeout{** loaded for the language `#1'. Using the pattern for}%
\typeout{** the default language instead.}%
\else
\language=\csname l@#1\endcsname
\fi
#2}}

\bibitem{laguna2019path}
G.~J. {Laguna} and S.~{Bhattacharya}, ``{Path planning with Incremental Roadmap
  Update for Visibility-based Target Tracking},'' in \emph{Proceedings of
  IEEE/RSJ International Conference on Intelligent Robots and Systems}, 2019,
  pp. 1159--1164.

\bibitem{bonatti2019towards}
R.~{Bonatti}, C.~{Ho}, W.~{Wang}, S.~{Choudhury}, and S.~{Scherer}, ``{Towards
  a Robust Aerial Cinematography Platform: Localizing and Tracking Moving
  Targets in Unstructured Environments},'' in \emph{Proceedings of IEEE/RSJ
  International Conference on Intelligent Robots and Systems}, 2019, pp.
  229--236.

\bibitem{fu2014robust}
C.~{Fu}, A.~{Carrio}, M.~A. {Olivares-Mendez}, R.~{Suarez-Fernandez}, and
  P.~{Campoy}, ``{Robust real-time vision-based aircraft tracking from Unmanned
  Aerial Vehicles},'' in \emph{Proceedings of IEEE International Conference on
  Robotics and Automation}, 2014, pp. 5441--5446.

\bibitem{henriques2015high}
J.~F. {Henriques}, R.~{Caseiro}, P.~{Martins}, and J.~{Batista}, ``{High-Speed
  Tracking with Kernelized Correlation Filters},'' \emph{IEEE Transactions on
  Pattern Analysis and Machine Intelligence}, vol.~37, no.~3, pp. 583--596,
  2015.

\bibitem{li2018learning}
F.~{Li}, C.~{Tian}, W.~{Zuo}, L.~{Zhang}, and M.~{Yang}, ``{Learning
  Spatial-Temporal Regularized Correlation Filters for Visual Tracking},'' in
  \emph{Proceedings of IEEE/CVF Conference on Computer Vision and Pattern
  Recognition}, 2018, pp. 4904--4913.

\bibitem{li2020autotrack}
Y.~Li, C.~Fu, F.~Ding, Z.~Huang, and G.~Lu, ``Autotrack: Towards
  high-performance visual tracking for uav with automatic spatio-temporal
  regularization,'' in \emph{Proceedings of the IEEE/CVF Conference on Computer
  Vision and Pattern Recognition}, 2020, pp. 11\,923--11\,932.

\bibitem{bertinetto2016fully}
L.~Bertinetto, J.~Valmadre, J.~F. Henriques, A.~Vedaldi, and P.~H. Torr,
  ``Fully-convolutional siamese networks for object tracking,'' in
  \emph{Proceedings of European Conference on Computer Vision}.\hskip 1em plus
  0.5em minus 0.4em\relax Springer, 2016, pp. 850--865.

\bibitem{valmadre2017end}
J.~{Valmadre}, L.~{Bertinetto}, J.~{Henriques}, A.~{Vedaldi}, and P.~H.~S.
  {Torr}, ``{End-to-End Representation Learning for Correlation Filter Based
  Tracking},'' in \emph{Proceedings of IEEE Conference on Computer Vision and
  Pattern Recognition}, 2017, pp. 5000--5008.

\bibitem{guo2017learning}
Q.~{Guo}, W.~{Feng}, C.~{Zhou}, R.~{Huang}, L.~{Wan}, and S.~{Wang},
  ``{Learning Dynamic Siamese Network for Visual Object Tracking},'' in
  \emph{Proceedings of IEEE International Conference on Computer Vision}, 2017,
  pp. 1781--1789.

\bibitem{grabner2006real}
H.~Grabner, M.~Grabner, and H.~Bischof, ``Real-time tracking via on-line
  boosting.'' in \emph{Proceedings of British Machine Vision Conference},
  vol.~1, no.~5, 2006, pp. 1--6.

\bibitem{babenko2011robust}
B.~{Babenko}, M.~{Yang}, and S.~{Belongie}, ``{Robust Object Tracking with
  Online Multiple Instance Learning},'' \emph{IEEE Transactions on Pattern
  Analysis and Machine Intelligence}, vol.~33, no.~8, pp. 1619--1632, 2011.

\bibitem{hare2016struck}
S.~{Hare}, S.~{Golodetz}, A.~{Saffari}, V.~{Vineet}, M.~{Cheng}, S.~L. {Hicks},
  and P.~H.~S. {Torr}, ``{Struck: Structured Output Tracking with Kernels},''
  \emph{IEEE Transactions on Pattern Analysis and Machine Intelligence},
  vol.~38, no.~10, pp. 2096--2109, 2016.

\bibitem{danelljan2017eco}
M.~{Danelljan}, G.~{Bhat}, F.~S. {Khan}, and M.~{Felsberg}, ``{ECO: Efficient
  Convolution Operators for Tracking},'' in \emph{Proceedings of IEEE
  Conference on Computer Vision and Pattern Recognition}, 2017, pp. 6931--6939.

\bibitem{Li2020ICRA}
F.~Li, C.~Fu, F.~Lin, Y.~Li, and P.~Lu, ``Training-set distillation for
  real-time uav object tracking,'' in \emph{Proceedings of IEEE International
  Conference on Robotics and Automation}, 2020, pp. 1--8.

\bibitem{bolme2010visual}
D.~S. Bolme, J.~R. Beveridge, B.~A. Draper, and Y.~M. Lui, ``Visual object
  tracking using adaptive correlation filters,'' in \emph{Proceedings of IEEE
  Computer Society Conference on Computer Vision and Pattern Recognition},
  2010, pp. 2544--2550.

\bibitem{danelljan2014adaptive}
M.~{Danelljan}, F.~S. {Khan}, M.~{Felsberg}, and J.~v.~d. {Weijer}, ``{Adaptive
  Color Attributes for Real-Time Visual Tracking},'' in \emph{Proceedings of
  IEEE Conference on Computer Vision and Pattern Recognition}, 2014, pp.
  1090--1097.

\bibitem{ma2015hierarchical}
C.~{Ma}, J.~{Huang}, X.~{Yang}, and M.~{Yang}, ``{Hierarchical Convolutional
  Features for Visual Tracking},'' in \emph{Proceedings of IEEE International
  Conference on Computer Vision}, 2015, pp. 3074--3082.

\bibitem{danelljan2015learning}
M.~{Danelljan}, G.~{H{\"a}ger}, F.~S. {Khan}, and M.~{Felsberg}, ``{Learning
  Spatially Regularized Correlation Filters for Visual Tracking},'' in
  \emph{Proceedings of IEEE International Conference on Computer Vision}, 2015,
  pp. 4310--4318.

\bibitem{dai2019visual}
K.~{Dai}, D.~{Wang}, H.~{Lu}, C.~{Sun}, and J.~{Li}, ``{Visual Tracking via
  Adaptive Spatially-Regularized Correlation Filters},'' in \emph{Proceedings
  of IEEE/CVF Conference on Computer Vision and Pattern Recognition}, 2019, pp.
  4665--4674.

\bibitem{danelljan2016adaptive}
M.~{Danelljan}, G.~{H{\"a}gerger}, F.~S. {Khan}, and M.~{Felsberg}, ``{Adaptive
  Decontamination of the Training Set: A Unified Formulation for Discriminative
  Visual Tracking},'' in \emph{Proceedings of IEEE Conference on Computer
  Vision and Pattern Recognition}, 2016, pp. 1430--1438.

\bibitem{danelljan2016beyond}
M.~Danelljan, A.~Robinson, F.~S. Khan, and M.~Felsberg, ``Beyond correlation
  filters: Learning continuous convolution operators for visual tracking,'' in
  \emph{Proceedings of European Conference on Computer Vision}.\hskip 1em plus
  0.5em minus 0.4em\relax Springer, 2016, pp. 472--488.

\bibitem{danelljan2017discriminative}
M.~{Danelljan}, G.~{H{\"a}ger}, F.~S. {Khan}, and M.~{Felsberg},
  ``{Discriminative Scale Space Tracking},'' \emph{IEEE Transactions on Pattern
  Analysis and Machine Intelligence}, vol.~39, no.~8, pp. 1561--1575, 2017.

\bibitem{li2014scale}
Y.~Li and J.~Zhu, ``A scale adaptive kernel correlation filter tracker with
  feature integration,'' in \emph{Proceedings of European Conference on
  Computer Vision}.\hskip 1em plus 0.5em minus 0.4em\relax Springer, 2014, pp.
  254--265.

\bibitem{mueller2017context}
M.~{Mueller}, N.~{Smith}, and B.~{Ghanem}, ``Context-aware correlation filter
  tracking,'' in \emph{Proceedings of IEEE Conference on Computer Vision and
  Pattern Recognition}, 2017, pp. 1387--1395.

\bibitem{huang2019learning}
Z.~Huang, C.~Fu, Y.~Li, F.~Lin, and P.~Lu, ``{Learning Aberrance Repressed
  Correlation Filters for Real-time UAV Tracking},'' in \emph{Proceedings of
  the IEEE International Conference on Computer Vision}, 2019, pp. 2891--2900.

\bibitem{zhu2018distractor}
Z.~Zhu, Q.~Wang, B.~Li, W.~Wu, J.~Yan, and W.~Hu, ``Distractor-aware siamese
  networks for visual object tracking,'' in \emph{Proceedings of the European
  Conference on Computer Vision}, 2018, pp. 101--117.

\bibitem{cheng2017an}
H.~{Cheng}, L.~{Lin}, Z.~{Zheng}, Y.~{Guan}, and Z.~{Liu}, ``An autonomous
  vision-based target tracking system for rotorcraft unmanned aerial
  vehicles,'' in \emph{Proceedings of IEEE/RSJ International Conference on
  Intelligent Robots and Systems}, 2017, pp. 1732--1738.

\bibitem{li2020keyfilter}
Y.~Li, C.~Fu, Z.~Huang, Y.~Zhang, and J.~Pan, ``Keyfilter-aware real-time uav
  object tracking,'' in \emph{Proceedings of IEEE International Conference on
  Robotics and Automation}, 2020, pp. 1--8.

\bibitem{li2020intermittent}
Y.~Li, C.~Fu, Z.~Huang, Y.~Zhang, and J.~Pan, ``Intermittent contextual
  learning for keyfilter-aware uav object tracking using deep convolutional
  feature,'' \emph{IEEE Transactions on Multimedia}, 2020.

\bibitem{Mueller2016ECCV}
M.~Mueller, N.~Smith, and B.~Ghanem, ``A benchmark and simulator for uav
  tracking,'' in \emph{Proceedings of European Conference on Computer
  Vision}.\hskip 1em plus 0.5em minus 0.4em\relax Springer, 2016, pp. 445--461.

\bibitem{Li2017AAAI}
S.~Li and D.-Y. Yeung, ``{Visual object tracking for unmanned aerial vehicles:
  A benchmark and new motion models},'' in \emph{Proceedings of Thirty-First
  AAAI Conference on Artificial Intelligence}, 2017, pp. 4140--4146.

\bibitem{Du2018ECCV}
D.~Du, Y.~Qi, H.~Yu, Y.~Yang, K.~Duan, G.~Li, W.~Zhang, Q.~Huang, and Q.~Tian,
  ``The unmanned aerial vehicle benchmark: object detection and tracking,'' in
  \emph{Proceedings of European Conference on Computer Vision}.\hskip 1em plus
  0.5em minus 0.4em\relax Springer, 2018, pp. 370--386.

\bibitem{galoogahi2017learning}
H.~K. {Galoogahi}, A.~{Fagg}, and S.~{Lucey}, ``{Learning Background-Aware
  Correlation Filters for Visual Tracking},'' in \emph{Proceedings of IEEE
  International Conference on Computer Vision}, 2017, pp. 1144--1152.

\bibitem{wang2018multi}
N.~{Wang}, W.~{Zhou}, Q.~{Tian}, R.~{Hong}, M.~{Wang}, and H.~{Li},
  ``{Multi-cue Correlation Filters for Robust Visual Tracking},'' in
  \emph{Proceedings of IEEE/CVF Conference on Computer Vision and Pattern
  Recognition}, 2018, pp. 4844--4853.

\bibitem{lukezic2017discriminative}
A.~{Luke{\v{z}}ic}, T.~{Voj{\'{i}}r}, L.~C. {Zajc}, J.~{Matas}, and
  M.~{Kristan}, ``{Discriminative Correlation Filter with Channel and Spatial
  Reliability},'' in \emph{Proceedings of IEEE Conference on Computer Vision
  and Pattern Recognition}, 2017, pp. 4847--4856.

\bibitem{wang2018kernel}
C.~Wang, L.~Zhang, L.~Xie, and J.~Yuan, ``Kernel cross-correlator,'' in
  \emph{Proceedings of Thirty-Second AAAI Conference on Artificial
  Intelligence}, 2018, pp. 4179--4186.

\bibitem{bertinetto2016staple}
L.~{Bertinetto}, J.~{Valmadre}, S.~{Golodetz}, O.~{Miksik}, and P.~H.~S.
  {Torr}, ``{Staple: Complementary Learners for Real-Time Tracking},'' in
  \emph{Proceedings of IEEE Conference on Computer Vision and Pattern
  Recognition}, 2016, pp. 1401--1409.

\bibitem{li2017integrating}
F.~{Li}, Y.~{Yao}, P.~{Li}, D.~{Zhang}, W.~{Zuo}, and M.~{Yang}, ``{Integrating
  Boundary and Center Correlation Filters for Visual Tracking with Aspect Ratio
  Variation},'' in \emph{Proceedings of IEEE International Conference on
  Computer Vision Workshops}, 2017, pp. 2001--2009.

\bibitem{li2019target}
X.~{Li}, C.~{Ma}, B.~{Wu}, Z.~{He}, and M.~{Yang}, ``{Target-Aware Deep
  Tracking},'' in \emph{Proceedings of IEEE/CVF Conference on Computer Vision
  and Pattern Recognition}, 2019, pp. 1369--1378.

\bibitem{wang2019unsupervised}
N.~{Wang}, Y.~{Song}, C.~{Ma}, W.~{Zhou}, W.~{Liu}, and H.~{Li},
  ``{Unsupervised Deep Tracking},'' in \emph{Proceedings of IEEE/CVF Conference
  on Computer Vision and Pattern Recognition}, 2019, pp. 1308--1317.

\bibitem{Dalal2005CVPR}
N.~Dalal and B.~Triggs, ``{Histograms of Oriented Gradients for Human
  Detection},'' in \emph{Proceedings of IEEE Computer Society Conference on
  Computer Vision and Pattern Recognition}, vol.~1, 2005, pp. 886--893.

\bibitem{wu2013online}
Y.~{Wu}, J.~{Lim}, and M.~{Yang}, ``{Online Object Tracking: A Benchmark},'' in
  \emph{Proceedings of IEEE Conference on Computer Vision and Pattern
  Recognition}, 2013, pp. 2411--2418.

\end{thebibliography}

\end{document}